\documentclass[%
paper=letter,%
pagesize,%
numbers=noendperiod,%
captions=nooneline,%
abstracton,%
DIV=14
]{scrartcl}

\usepackage[T1]{fontenc}%
\usepackage{lmodern}%
\usepackage[american]{babel}%
\usepackage{microtype}%

\usepackage[
hyperref,%
table%
]{xcolor}%

\usepackage{scrlayer-scrpage}%

\usepackage{eso-pic}%
\usepackage{rotating}%
\usepackage{amsmath}%
\usepackage{mathtools}%
\usepackage{amssymb}%
\usepackage{amsthm}%
\usepackage{etoolbox}%
\usepackage{bm}%
\usepackage{bbm}%
\usepackage{enumitem}%
\usepackage{graphicx}%
\usepackage{siunitx} %
\usepackage{grffile}%
\usepackage{tikz}%
\usetikzlibrary{shapes, decorations.pathreplacing}
\usepackage{wrapfig}%
\usepackage{tabularx}%
\usepackage{bbm}
\usepackage{siunitx}%
\usepackage{booktabs}%
\usepackage{multirow}%
\usepackage{fancyvrb}%
\usepackage{listings}%
\usepackage{csquotes}%
\usepackage[
style=authoryear,%
dashed=false,%
hyperref=true,%
useprefix=true,%
maxnames=2,%
maxbibnames=6,%
uniquename=false,%
seconds=true,%
date=iso,%
urldate=iso%
]{biblatex}%
\usepackage[
hypertexnames=false,%
setpagesize=false,%
pdfborder={0 0 0},%
pdfstartview=Fit,%
bookmarksopen=true,%
bookmarksnumbered=true%
]{hyperref}%
\xdefinecolor{lightgray}{RGB}{247, 247, 247}%
\xdefinecolor{semilightgray}{RGB}{240, 240, 240}%
\xdefinecolor{middlegray}{RGB}{127, 127, 127}%
\xdefinecolor{blue}{RGB}{58, 95, 205}%
\xdefinecolor{deepskyblue}{RGB}{0, 154, 205}%
\xdefinecolor{chocolate}{RGB}{205, 102, 29}%

\setkomafont{pageheadfoot}{\normalfont\normalcolor\sffamily}%
\setkomafont{pagenumber}{\normalfont\normalcolor\sffamily}%
\automark{section}%
\setkomafont{captionlabel}{\normalfont\normalcolor\sffamily\bfseries}%

\setlist{%
  align=left,%
  labelsep=*,%
  leftmargin=*,%
  topsep=1mm,%
  itemsep=0mm%
}
\newcommand*{\mysquare}{\rule[0.18em]{0.36em}{0.36em}}
\newcommand*{\mytriangle}{\raisebox{0.12em}{\resizebox{0.48em}{0.48em}{$\blacktriangleright$}}}
\newcommand*{\mybar}{\rule[0.32em]{0.62em}{0.08em}}
\newcommand*{\mydot}{\raisebox{0.14em}{\resizebox{0.44em}{!}{$\bullet$}}}
\setlist[itemize,1]{label={\mysquare\ }}%
\setlist[itemize,2]{label={\mytriangle\ }}%
\setlist[itemize,3]{label={\mybar\ }}%
\setlist[itemize,4]{label={\mydot\ }}%
\setlist[enumerate,1]{label=\arabic*)}%
\setlist[enumerate,2]{label=\arabic{enumi}.\arabic*)}%
\setlist[enumerate,3]{label=\arabic{enumi}.\arabic{enumii}.\arabic*)}%

\makeatletter
\newcommand\myisodate{\number\year-\ifcase\month\or 01\or 02\or 03\or 04\or 05\or 06\or 07\or 08\or 09\or 10\or 11\or 12\fi-\ifcase\day\or 01\or 02\or 03\or 04\or 05\or 06\or 07\or 08\or 09\or 10\or 11\or 12\or 13\or 14\or 15\or 16\or 17\or 18\or 19\or 20\or 21\or 22\or 23\or 24\or 25\or 26\or 27\or 28\or 29\or 30\or 31\fi}%
\makeatother
\newcommand*{\abstractnoindent}{}%
\let\abstractnoindent\abstract
\renewcommand*{\abstract}{\let\quotation\quote\let\endquotation\endquote
  \abstractnoindent}
\deffootnote[1em]{1em}{1em}{\textsuperscript{\thefootnotemark}}%
\lstdefinestyle{input}{
  backgroundcolor=\color{semilightgray},%
  commentstyle=\itshape\color{chocolate},%
  keywordstyle=\color{blue},%
  emphstyle=\color{blue},%
  stringstyle=\color{black},%
  numbers=left,%
  numbersep=4.8pt,%
  numberstyle=\color{darkgray!80}\tiny%
}
\lstdefinestyle{output}{
  backgroundcolor=\color{lightgray}%
}

\lstdefinestyle{Rstyle}{
  language=R,%
  keywords={function, if, else, switch, repeat, while, for, in, next, break},%
  otherkeywords={},%
  emph={TRUE, FALSE, NULL, NA, NaN, Inf}%
}
\expandafter\let\csname Sinput\endcsname\relax
\expandafter\let\csname endSinput\endcsname\relax
\expandafter\let\csname Soutput\endcsname\relax
\expandafter\let\csname endSoutput\endcsname\relax
\lstnewenvironment{Sinput}[1][]{%
  \lstset{style=input, style=Rstyle}
  #1%
}{\vspace{-0.25\baselineskip}}%
\lstnewenvironment{Soutput}[1][]{%
  \lstset{style=output, style=Rstyle}
  #1%
}{\vspace{-0.25\baselineskip}}%

\lstdefinestyle{LaTeXstyle}{
  language=[LaTeX]TeX,%
  texcs={},%
  otherkeywords={}%
}
\lstnewenvironment{LaTeXinput}[1][]{%
  \lstset{style=input, style=LaTeXstyle}
  #1%
}{\vspace{-0.25\baselineskip}}%
\lstnewenvironment{LaTeXoutput}[1][]{%
  \lstset{style=output, style=LaTeXstyle}
  #1%
}{\vspace{-0.25\baselineskip}}%

\lstdefinestyle{otherstyle}{
  language={},%
  otherkeywords={},%
  upquote=true%
}
\lstnewenvironment{otherinput}[1][]{%
  \lstset{style=input, style=otherstyle}
  #1%
}{\vspace{-0.25\baselineskip}}%
\lstnewenvironment{otheroutput}[1][]{%
  \lstset{style=output, style=otherstyle}
  #1%
}{\vspace{-0.25\baselineskip}}%
\DeclareNameAlias{sortname}{last-first}%
\DefineBibliographyExtras{american}{\DeclareQuotePunctuation{}}%
\renewbibmacro*{volume+number+eid}{%
  \setunit*{\addcomma\space}%
  \printfield{volume}%
  \printfield{number}}
\DeclareFieldFormat*{number}{(#1)}
\DeclareFieldFormat*{title}{#1}%
\DeclareFieldFormat{doi}{%
  \ifhyperref
  {\href{http://dx.doi.org/#1}{\nolinkurl{doi:#1}}}%
  {\nolinkurl{doi:#1}}}%
\renewbibmacro*{in:}{}%
\DeclareFieldFormat{isbn}{ISBN #1}%
\DeclareFieldFormat{pages}{#1}%
\DeclareFieldFormat{url}{\url{#1}}%
\DeclareFieldFormat{urldate}{\mkbibparens{#1}}%
\renewcommand*{\cite}[2][]{\textcite[#1]{#2}}%

\newif\ifstarttheorem
\newtheoremstyle{mythmstyle}%
{0.5em}%
{0.5em}%
{}%
{}%
{\sffamily\bfseries\global\starttheoremtrue}%
{}%
{\newline}%
{\thmname{#1}\ \thmnumber{#2}\ \thmnote{(#3)}}%
\theoremstyle{mythmstyle}%
\newtheorem{definition}{Definition}[section]%

\newtheorem{algorithm}[definition]{Algorithm}

\makeatletter
\preto\itemize{%
  \if@inlabel
  \ifstarttheorem
  \mbox{}\par\nobreak\vskip\glueexpr-\parskip-\baselineskip+0.25em\relax\hrule\@height\z@
  \fi%
  \fi%
  \global\starttheoremfalse%
  \def\tempa{proof}%
  \ifx\tempa\mycurrenvir
  \ifstarttheorem
  \mbox{}\par\nobreak\vskip\glueexpr-\parskip-\baselineskip+0.25em\relax\hrule\@height\z@
  \fi%
  \fi%
  \global\starttheoremfalse%
}
\preto\enditemize{\global\starttheoremfalse}
\makeatother

\makeatletter
\preto\enumerate{%
  \if@inlabel
  \ifstarttheorem
  \mbox{}\par\nobreak\vskip\glueexpr-\parskip-\baselineskip+0.25em\relax\hrule\@height\z@
  \fi%
  \fi%
  \global\starttheoremfalse%
  \def\tempa{proof}%
  \ifx\tempa\mycurrenvir
  \ifstarttheorem
  \mbox{}\par\nobreak\vskip\glueexpr-\parskip-\baselineskip+0.25em\relax\hrule\@height\z@
  \fi%
  \fi%
  \global\starttheoremfalse%
}
\preto\endenumerate{\global\starttheoremfalse}
\makeatother

\newcommand{\ou}[3]{%
  \mathrel{%
    \vcenter{\offinterlineskip
      \ialign{##\cr$#1$\cr\noalign{\kern-#3}$#2$\cr}%
    }%
  }%
}

\newcommand*{\T}{^{\top}}

\newcommand*{\IN}{\mathbbm{N}}

\newcommand*{\IR}{\mathbbm{R}}

\newcommand*{\N}{\operatorname{N}}

\newcommand*{\I}{\mathbbm{1}}
\newcommand*{\rd}{\mathrm{d}}

\renewcommand*{\P}{\mathbbm{P}}
\newcommand*{\E}{\mathbbm{E}}

\newcommand*{\AMSE}{\operatorname{AMSE}}

\newcommand*{\CvM}{\operatorname{CvM}}

\newcommand*{\R}{\textsf{R}}

\newcommand*{\ntrn}{n_{\text{trn}}}
\newcommand*{\ntst}{n_{\text{tst}}}

\newcommand*{\nepo}{n_{\text{epo}}}
\newcommand*{\ngen}{n_{\text{gen}}}

\newcommand*{\nrep}{n_{\text{rep}}}

\begin{document}
\thispagestyle{plain}
\begin{center}
  \sffamily
  {\bfseries\LARGE %
    RafterNet: Probabilistic Predictions\\[2mm] in Multi-Response Regression\par}
  \bigskip\smallskip
  {\Large Marius Hofert\footnote{Department of Statistics and Actuarial Science, University of
      Waterloo, 200 University Avenue West, Waterloo, ON, N2L
      3G1,
      \href{mailto:marius.hofert@uwaterloo.ca}{\nolinkurl{marius.hofert@uwaterloo.ca}}. The
      author acknowledges support from NSERC (Grant RGPIN-2020-04897).},
    Avinash Prasad\footnote{Department of Statistics and Actuarial Science, University of
      Waterloo, 200 University Avenue West, Waterloo, ON, N2L
      3G1,
      \href{mailto:a2prasad@uwaterloo.ca}{\nolinkurl{a2prasad@uwaterloo.ca}}. The author acknowledges support from Fin-ML CREATE scholarship.},
    Mu Zhu\footnote{Department of Statistics and Actuarial Science, University of
      Waterloo, 200 University Avenue West, Waterloo, ON, N2L
      3G1,
      \href{mailto:mu.zhu@uwaterloo.ca}{\nolinkurl{mu.zhu@uwaterloo.ca}}. The
      author acknowledges support from NSERC (RGPIN-2016-03876).}
    \par\bigskip
    \myisodate\par}
\end{center}
\par\smallskip
\begin{abstract}
  A fully nonparametric approach for making probabilistic predictions in
  multi-response regression problems is introduced. Random forests are used as
  marginal models for each response variable and, as novel contribution of the
  present work, the dependence between the multiple response variables is
  modeled by a generative neural network. This combined modeling approach of
  random forests, corresponding empirical marginal residual distributions and a
  generative neural network is referred to as RafterNet. Multiple datasets serve
  as examples to demonstrate the flexibility of the approach and its impact for
  making probabilistic forecasts.
\end{abstract}
\minisec{Keywords}
Multi-response regression,
learning distributions,
probabilistic forecasts,
random forests,
copulas,
generative neural networks.
\minisec{MSC2010}
62H99, 65C60, 62J99, 62E17 %

\section{Introduction}
We consider a fairly general class of problems, where the joint distribution of
a $d$-dimensional random vector $\bm{X}_k=(X_{k,1},\dots,X_{k,d})$ allows for
the Sklar decomposition \parencite{sklar1959},
\begin{align}
  F_{\bm{X}_k}(\bm{x}_k) &= C(F_{X_{k,1}}(x_{k,1}),\dots,F_{X_{k,d}}(x_{k,d})), \label{eq:specialSklar}
\end{align}
in which the copula $C$ \parencite{nelsen2006,embrechtsmcneilstraumann2002}
remains the same across all $k$ and satisfies the ``simplifying
  assumption'' in the sense of \cite{cotegenestomelka2019}, but the marginal distributions
$F_{X_{k,1}},\dots,F_{X_{k,d}}$ can depend on a (vector) covariate, say,
$\bm{z}_k$, and hence vary with $k$.

\subsection{Background}
Since \cite{song2000,oakesritz2000} first presented multi-response regression
modeling using copulas, it has been explored in a few different contexts,
including insurance applications
\parencite{freeswang2005,freeswang2006,freesvaldez2008, cotegenestomelka2019}
and survival analysis
\parencite{helawless2005,barrigalouzadanetoortegacancho2010}. Typically, each
marginal distribution is assumed to follow a parametric model,
\begin{align}
  F_{X_{k,j}}(x_{k,j}) \equiv F_j(x_{k,j}; \theta_j(\bm{z}_k)),\quad j=1,\dots,d, \label{eq:parametric_marg}
\end{align}
with parameter $\theta_j(\bm{z}_k)$ depending on the covariate $\bm{z}_k$, for
example specified by a generalized linear model (GLM) with
$\theta_j(\bm{z}_k)=\eta_j(\bm{\beta}\T_j\bm{z}_k)$, where $\eta_j$ is a
pre-specified link function specific to the parametric family of $F_j$. Then, a
parametric copula model (for example normal, $t$, Frank, Gumbel, and so on) is
chosen as $C$; see \cite{gijbelsomelkaveraverbeke2015, cotegenestomelka2019} for
investigations into the estimation and selection of $C$ in copula-based
regression setups.

In any dimension $j$, the function $\theta_j(\cdot)$ can be estimated from training
data $\{(X_{k,j}, \bm{z}_k)\}_{k=1}^{\ntrn}$. Afterwards, one can apply the
probability integral transform to each training observation
\begin{align}
  \widehat{U}_{k,j}&=F_j(X_{k,j}; \widehat{\theta}_j(\bm{z}_k)), \quad k=1,\dots,\ntrn, \label{eq:deMargin_param}
\end{align}
and estimate the copula $C$ from the transformed sample,
\begin{align*}
  \{\widehat{\bm{U}}_k = (\widehat{U}_{k,1},\dots,\widehat{U}_{k,d})\}_{k=1}^{\ntrn}.
\end{align*}

The goal is to make probabilistic predictions for any $k$, either a training
observation ($k\leq \ntrn$) or a future observation ($k>\ntrn$). This can be
done by first generating a sample of size $\ngen$,
\begin{align*}
  \{\bm{U}^{(i)}_{k}=(U^{(i)}_{k,1},\dots,U^{(i)}_{k,d})\}_{i=1}^{\ngen},
\end{align*}
from the estimated copula $\widehat{C}$, and then letting
\begin{align}
  \widehat{X}^{(i)}_{k,j}= F^{-1}_j(U^{(i)}_{k,j}; \widehat{\theta}_j(\bm{z}_{k})),\quad i=1,\dots,\ngen. \label{eq:addMargin_param}
\end{align}
The resulting collection
\begin{align*}
  \{\widehat{\bm{X}}_k^{(i)} = (\widehat{X}^{(i)}_{k,1},\dots,\widehat{X}^{(i)}_{k,d})\}_{i=1}^{\ngen}
\end{align*}
is an empirical predictive distribution for $\bm{X}_k$. We can then make
probabilistic forecasts such as predicting
$\P(X_{k,j}>c_j,\ X_{k,\ell}>c_{\ell})$ by
$(1/\ngen)\sum_{i=1}^{\ngen} \I(\widehat{X}^{(i)}_{k,j}>c_j,\
\widehat{X}^{(i)}_{k,\ell}>c_{\ell})$, something that point
predictions/forecasts are incapable of; see Appendix~\ref{sec:R:implementation}
for such an example based on our fully nonparametric approach to be
detailed below. For these types of predictions, correctly capturing the
dependence structure $C$ is critical.

\subsection{Our contribution}
The aforementioned classic, and fully parametric, approach will be
illustrated using one dataset later in Appendix~\ref{appdx:marg} as a
comparison.  For the main part of this article, however, we will
  propose a fully nonparametric approach.  For the marginal models
$F_{X_k,1},\dots,F_{X_k,d}$, this objective is easy to achieve. Instead of the
parametric approach \eqref{eq:parametric_marg}, we model the mean in each
dimension $j$ as a function of the covariate,
\begin{align*}
\E(X_{k,j})=\theta_j(\bm{z}_k),
\end{align*}
fitted with a random forest \parencite{breiman2001}, and the distribution of the
ensuing residual,
\begin{align*}
X_{k,j}-\E(X_{k,j})\sim F_j,
\end{align*}
with its empirical (rather than a specific parametric) distribution function.

Our key contribution, and the main focus of this article, is to model
the joint distribution of the transformed variables
\begin{align*}
  \bigl(F_1(X_{k,1}-\E(X_{k,1})), \dots, F_d(X_{k,d}-\E(X_{k,d}))\bigr) \sim C
\end{align*}
by a generative neural network. Instead of fitting and then sampling from
a parametric copula for $C$, we train a neural network to directly provide us with samples
from an estimate of $C$.

\begin{figure}
  \centering
  \includegraphics[width=0.7\textwidth]{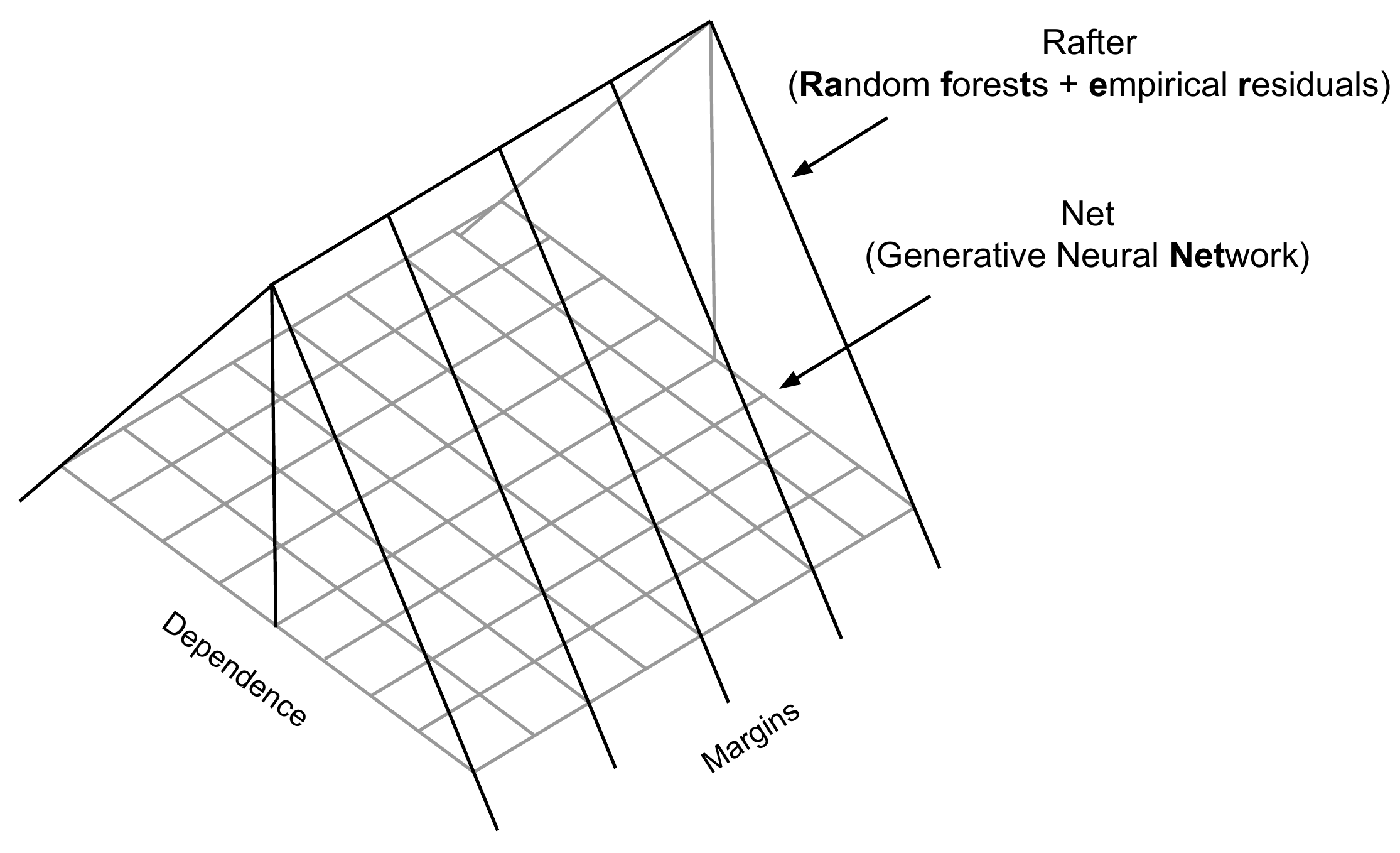}%
  \caption{Illustration of the idea behind RafterNet. Random forests provide
    marginal models, the empirical residuals of which are then used as input to
    a neural network that models the dependence
    structure.}\label{fig:rafternet}
\end{figure}

We have found this approach to be quite powerful in practice, to a large extent
due to its considerable flexibility at all three levels: $\theta_j(\bm{z}_k)$ for
$\E(X_{k,j})$, $F_j$ for $X_{k,j}-\E(X_{k,j})$, $j=1,\dots,d$, and a neural network for
$\bigl(F_1(X_{k,1}-\E(X_{k,1})),\dots,F_d(X_{k,d}-\E(X_{k,d}))\bigr)$.
Therefore, we refer to our model as RafterNet, for ``\underline{ra}ndom
\underline{f}ores\underline{t}s + \underline{e}mpirical \underline{r}esiduals +
generative neural \underline{net}work''. The English word ``rafter'' means ``one
of several internal beams extending from the eaves to the peak of a roof and
constituting its framework''. The analogy is especially apt here: The $j$th beam
represents the $j$th marginal model, and these marginal models
are then connected by the neural network which learns the dependence
across all dimensions $j=1,\dots,d$; see Figure~\ref{fig:rafternet} for a
schematic illustration.

Any nonparametric regression technique can be used to model
$\theta_j(\cdot)$; the random forest is merely being used as a generic choice
which has the reputation of being both relatively robust and easy to apply. In
Appendix~\ref{appdx:marg}, we illustrate with an example that, even when using
classic GLMs as marginal models, it can still be beneficial to model the
dependence across all dimensions with a neural network, as opposed to a
parametric copula.

\section{The RafterNet}\label{sec:main}
\subsection{Modeling and probabilistic prediction}
Fitting a random forest to estimate the marginal regression function
$\theta_j(\cdot)$ is straightforward. For any observation $k$, let
$\widehat{R}_{k,j} = X_{k,j} - \widehat{\theta}_j(\bm{z}_k)$ denote the
realized residual in the $j$th coordinate after having removed the estimated
effect of the covariate. As typical in copula modeling, we use
\begin{align*}
  \widehat{F}_{j}(t)=\frac{1}{\ntrn+1}\sum_{\ell=1}^{\ntrn} \I(\widehat{R}_{\ell,j} \leq t)
\end{align*}
as the empirical distribution of $\widehat{R}_{1,j},\dots,\widehat{R}_{\ntrn,j}$.
Thus, for us, Equation~\eqref{eq:deMargin_param} corresponds to
\begin{align}
  \widehat{U}_{k,j} = \widehat{F}_j(X_{k,j} - \widehat{\theta}_j(\bm{z}_k)). \tag{\ref{eq:deMargin_param}'}\label{eq:deMargin_nonparam}
\end{align}
By letting $\widehat{\bm{U}}_k = (\widehat{U}_{k,1},\dots,\widehat{U}_{k,d})$
for all $k=1,\dots,\ntrn$, we then train a neural network $G$ in such
a way that, given any sample $\{\bm{V}_{\ell}\}_{\ell=1}^{\ngen}$ from a
``simple'' distribution (for example the uniform or the independent standard
normal), the two samples
\begin{align*}
  \{\widehat{\bm{U}}_k\}_{k=1}^{\ntrn} \quad\text{and}\quad \{\bm{U}_{\ell} = G(\bm{V}_{\ell})\}_{\ell=1}^{\ngen}
\end{align*}
follow approximately the same distribution; more details about this step are given in Section~\ref{sec:gnnTraining} below.
For any given $k$, this allows us to generate samples
\begin{align*}
  \{\bm{U}_{k}^{(i)} = G(\bm{V}_{k}^{(i)})\}_{i=1}^{\ngen}
\end{align*}
from an \emph{implicit} estimate of $C$, say, $\widehat{C}$, without making any
parametric assumptions about $C$.  Finally, by utilizing the quantile function
$\widehat{F}^{-1}_{j}$ of $\widehat{F}_j$, Equation~\eqref{eq:addMargin_param}
becomes
\begin{align}
  \widehat{X}^{(i)}_{k,j} = \widehat{F}^{-1}_j(U^{(i)}_{k,j}) + \widehat{\theta}_j(\bm{z}_k). \tag{\ref{eq:addMargin_param}'}\label{eq:addMargin_nonparam}
\end{align}

Algorithm~\ref{tab:rafternet_predict} summarizes the main steps used by our suggested
RafterNet model to make probabilistic predictions given a covariate $\bm{z}$.
\begin{algorithm}[Using the RafterNet to make probabilistic predictions given a covariate]\label{tab:rafternet_predict}
  \begin{enumerate}
  \item From the trained neural network $G$, generate $\bm{U}^{(i)}=\big(U^{(i)}_1,\dots,U^{(i)}_d\big) \sim \widehat{C}$, $i=1,\dots,\ngen$.
  \item For all $i=1,\dots,\ngen$ and $j=1,\dots,d$, let $\widehat{X}^{(i)}_{j}=\widehat{F}^{-1}_j(U^{(i)}_{j})+\widehat{\theta}_j(\bm{z})$.
  \item Return
    $\{(\widehat{X}^{(i)}_1,\dots,\widehat{X}^{(i)}_d)\}_{i=1}^{\ngen}$ as a
    sample from the empirical predictive distribution.
  \end{enumerate}
\end{algorithm}

\subsection{The optimization problem}\label{sec:gnnTraining}
We now briefly describe how to train a neural network
$G$ capable of generating samples of approximately the same distribution as
a given training sample; for us, the latter is $\{\widehat{\bm{U}}_k\}_{k=1}^{\ntrn}$. We use
a technique introduced independently by \cite{liswerskyzemel2015} and \cite{dziugaiteroyghahramani2015}.

Let $\mathcal{G}$ denote a family of feedforward neural networks $G$ with a
pre-determined architecture, and let $K$ be a kernel function. Given
$\{\bm{V}_{\ell} \in \IR^{d'}\}_{\ell=1}^{\ngen}$ from a ``simple'' distribution (for example
the uniform or the independent standard normal), we solve the optimization
problem
\begin{align}
\underset{G\in\mathcal{G}}\min\Biggl\{\,
  \frac{1}{\ntrn^2}\sum_{k=1}^{\ntrn}\sum_{k'=1}^{\ntrn}   K(\widehat{\bm{U}}_k,\widehat{\bm{U}}_{k'})
  -\frac{2}{\ntrn\ngen} \sum_{k=1}^{\ntrn}\sum_{\ell=1}^{\ngen} K(\widehat{\bm{U}}_k,G(\bm{V}_{\ell}))+\frac{1}{\ngen^2}\sum_{\ell=1}^{\ngen}\sum_{\ell'=1}^{\ngen} K(G(\bm{V}_{\ell}),G(\bm{V}_{\ell'}))\Biggr\}\label{eq:opt_K}
\end{align}
by stochastic gradient descent. The first term in \eqref{eq:opt_K} does not depend on $G$, so it does not have any direct impact on the optimization problem itself, but retaining it in the equation will make the optimization problem easier to understand.

In particular, the kernel function computes inner products in an implicit feature space \parencite{mercer1909} in the sense that $K(\bm{u},\bm{v})=\varphi(\bm{u})\T\varphi(\bm{v})$ for a feature map $\varphi$. Therefore, \eqref{eq:opt_K} is equivalent to
\begin{align}
\underset{G \in \mathcal{G}}{\min}\,\left\|\frac{1}{\ntrn}\sum_{k=1}^{\ntrn} \varphi(\widehat{\bm{U}}_k)-
       \frac{1}{\ngen}\sum_{\ell=1}^{\ngen} \varphi(G(\bm{V}_{\ell})) \right\|^2. \label{eq:opt_phi}
\end{align}
The feature map implicitly specified by the Gaussian kernel (here, with bandwidth parameter $h$),
\begin{align}
  K(\bm{u},\bm{v})=\exp\left(-\frac{\|\bm{u}-\bm{v}\|^2}{h}\right),\label{eq:GaussKer}
\end{align}
is an infinite-dimensional vector function $\varphi:\IR^d \to \IR^{\IN}$. %
With such a choice, the two terms in \eqref{eq:opt_phi}
will contain all empirical moments of $\{\widehat{\bm{U}}_k\}_{k=1}^{\ntrn}$ and
$\{G(\bm{V}_{\ell})\}_{\ell=1}^{\ngen}$, respectively. It is in this sense that
the solution $G$ can generate samples that ``match'' the training sample in
distribution.

In practice, we always set $d'=d$. We also follow the suggestion of
\cite{liswerskyzemel2015} and use a mixture of Gaussian kernels with different
bandwidths (instead of a single Gaussian kernel) in order to avoid having to
select an ``optimal'' bandwidth parameter; this is particularly convenient as
the output of our neural network always lies in the unit hypercube.  For an
investigation into these neural networks, an application to generating
quasi-random numbers from complex dependence structures, and details about
training these neural networks, see \cite{hofertprasadzhu2021}.

\subsection{Remarks}%
\label{sec:vae}

One may ask why we have chosen the technique of \cite{liswerskyzemel2015} and
\cite{dziugaiteroyghahramani2015}, instead of some other techniques such as
variational auto-encoders (VAEs), for training generative neural networks. The
short, and not-so-surprising, answer is that this technique works while others
do not. It is true that learning a VAE \parencite{kingma2013} will also allow
us to generate from $\{\bm{V}_{\ell} \in \IR^{d'}\}_{\ell=1}^{\ngen}$ a sample
$\{G(\bm{V}_{\ell})\in\IR^d\}_{\ell=1}^{\ngen}$ that follows a certain target
distribution, but VAEs make the explicit assumption that this target
distribution \emph{is concentrated around a smooth manifold} in $\IR^d$, usually
having a much lower intrinsic dimensionality than $d$. The main effort of the
VAE is to learn this unknown manifold from training data. For our learning
problem, however, this crucial assumption does \emph{not} apply. Our target
distributions are copulas in $\IR^d$, and they usually do \emph{not} concentrate
around some lower-dimensional manifold. As a result, VAEs --- and other
techniques which also rely heavily on this ``manifold assumption'' --- simply
cannot be used to learn what we really want to learn here. We shall present some
empirical evidence to this effect in Appendix~\ref{appdx:vae}.

\section{Examples}
\label{sec:examples}
In this section we apply RafterNet to a variety of multi-response regression
examples. We focus on our key innovation, the nonparametric modeling of $C$, and
compare our approach based on a neural network with the conventional approach of using a
parametric copula. For a fair comparison, we also combine the latter with random
forests as models for each $\E(X_j)$. For ease of comparison with RafterNets, we
refer to the latter model based on a parametric copula as ``RafterCop''. Both
RafterCops and RafterNets therefore share the same marginal models and that the
residuals $X_j-\E(X_j)$, $j=1,\dots,d$, are modeled empirically, but we replace
the neural network (the ``Net'' part) with a conventional copula (the ``Cop''
part) when modeling the joint distribution of
$\bigl(F_1(X_1-\E(X_1)),\dots,F_d(X_d-\E(X_d))\bigr)$. A summary of our findings
across all datasets is provided in Section~\ref{sec:examples:results}.

\subsection{Datasets}\label{sec:datasets}
First, we consider a demographic dataset containing observations of height and
weight of the !Kung San people in Botswana collected by \cite{howell2009}. The
exact dataset we use can be found on the webpage by \cite{mcelreath2020} under the
name \texttt{Howell1.csv}; see also Appendix~\ref{sec:R:implementation}. We are
interested in modeling the distribution of the height and weight of individuals
conditional on their age and sex. All models we consider are trained on
$\ntrn=444$ randomly selected individuals and the remaining $\ntst=100$ samples
serve as test data to evaluate the models.

Second, we consider a dataset with results from the 2019 Ironman World
Championship held in Hawaii. It can be downloaded from \cite{ironman2019}. Using
these data, we aim to model the joint distribution of swimming, biking and
running times of all competitors conditional on their ``region of
representation'' (or continent) and ``race category'' (indicating the
professional status, age group and sex of each competitor). We use results
from $\ntrn=1959$ randomly chosen athletes to train all models we consider and
the remaining $\ntst=300$ observations to evaluate them.

Third, we consider a dataset obtained from the National Education Longitudinal
Study (NELS) of 1988 \parencite{curtiningelswuheuer2002}. It can be downloaded
from \cite{necs2021}; on this website, follow the link ``1988-00'', then
``Download'', under ``Statistical Software Formats'' use ``R'' and finally
download the dataset and its explanations from the two appearing links. The NELS
was a major study in the US that measured the educational achievement and growth
of a nationally representative sample of middle school students (from 1052
public and private schools) along with numerous factors that could potentially
impact a student's academic performance. In this example, we are particularly
interested in modeling the joint distribution of the standardized scores (in the
base year 1988) for mathematics, science, reading comprehension and social
studies, conditional on 10 covariates, which are sex, race,
  socioeconomic status, minority, family size, family composition, school size,
  urbanicity, school type and student--teacher ratio. We use
$\ntrn=9888$ randomly chosen observations to train RafterNet and RafterCop
models and the remaining $\ntst=1000$ observations to assess the quality of
probabilistic predictions produced by these models.

Fourth, we consider a dataset extracted from the air quality system database of
the Environmental Protection Agency (EPA), see \parencite{EPA2021}, with the
help of the \R\ package \texttt{RAQSAPI} that provides an API to the EPA
database. The dataset contains air sample data collected by state, local, tribal
and federal air pollution control agencies from various monitoring stations
across the US. Because of missing data, we aggregate the data across all
monitoring sites (and over potentially multiple measurement devices per
site). We have $d=8$ variables of interest, which are the levels of Carbon
Monoxide (CO; measured in parts per millions (ppm)), Nitrogen Dioxide (NO$_2$;
measured in parts per billion (ppb)), Oxides of Nitrogen (NO$_x$; ppb), Ozone
(O$_3$; ppm), Sulfur Dioxide (SO$_2$; ppb), Carbon Dioxide (CO$_2$; ppm),
particulate matter in the air with diameter of 10 microns or less (PM$_{10}$;
measured in micrograms per cubic meter ($\mu\text{g}/\text{m}^3$)) and
particulate matter in the air with diameter of 2.5 microns or less (PM$_{2.5}$;
$\mu\text{g}/\text{m}^3$). We are then interested in modeling the distribution
of these eight air pollutants conditional on six covariates, which are
barometric pressure (measured in millibar), temperature (measured in degrees
Fahrenheit), relative humidity (in percent), wind speed (measured in knots),
rain (measured as a total in inches over a 24 hour period) and the day of the
week. We use $\ntrn=1322$ randomly chosen observations for training our models
and the remaining $\ntst=300$ observations for assessing probabilistic
predictions generated from the models.

\subsection{Results}\label{sec:examples:results}
We created a variety of RafterCops, each using a different copula (such as
normal, $t$, vine, Frank, Gumbel, empirical and empirical beta) to model $C$. We
compared these RafterCops with a variety of RafterNets, each using a different
neural network architecture, referred to by the notation
``$\text{G}_{h}^{\ell\text{x}}$'', where $\ell$ is the number of hidden layers
and $h$ the number of neurons per hidden layer.

We assess two aspects of the out-of-sample performance of RafterCops and
RafterNets. First, we evaluate how close generated samples from the neural
networks and copulas are to the underlying dependence of response observations
$\bm{X}_k$ in the test dataset.  To do so, we use a two-sample Cram\'{e}r-von-Mises type
test statistic \parencite{remillardscaillet2009} that is averaged over
$\nrep$-many replications and is defined by
\begin{align}
  \text{ACvM} = \frac{1}{\nrep}\sum_{i=1}^{\nrep} \Biggl(\frac{1}{\sqrt{\frac{1}{\ntst}+\frac{1}{\ngen}}}\int_{[0,1]^d} \bigl(C_{\ntst}(\bm{u})-C^{(i)}_{\ngen}(\bm{u})\bigr)^2\,\mathrm{d}\bm{u}\Biggr), \label{eq:acvm}
\end{align}
where $C_{\ntst}$ is the empirical copula of the $\ntst$ observations in the
test dataset and $C^{(i)}_{\ngen}$ is the empirical copula of the $\ngen$
samples generated from either a neural network or a copula in replication
$i$. Let $\mathcal{T}$ denote the set of indices $k$ for which
$(\bm{X}_k,\bm{z}_k)$ is in the test dataset. We can extract the $C_{\ntst}$ of
the test dataset observations by computing
$\widehat{\bm{U}}_{k}=\bigl(\widehat{F}_1(X_{k,1} -
\widehat{\theta}_1(\bm{z}_k)),\dots,\widehat{F}_d(X_{k,d} -
\widehat{\theta}_d(\bm{z}_k))\bigr)$ for $k\in\mathcal{T}$, where, for
$j=1,\dots,d$, $\hat{\theta}_{j}$ and $\hat{F}_j$ are the marginal fitted random
forests and empirical distributions based on the training data
$\{(X_{k,j}, \bm{z}_k)\}_{k=1}^{\ntrn}$. In our experiments, we use $\nrep=25$
replications to compute the ACvM metric~\eqref{eq:acvm}.

Second, we evaluate the quality of probabilistic predictions produced by
RafterCops and RafterNets using the average mean squared error over all test observations
\begin{align*}
  \text{AMSE} = \frac{1}{|\mathcal{T}|} \sum_{k\in\mathcal{T}}
  \Biggl(\frac{1}{\ngen}\sum_{i=1}^{\ngen} \|\widehat{\bm{X}}_k^{(i)}-\bm{X}_k\|^2\Biggr).
\end{align*}
We use $\ngen=1000$ samples when computing the AMSE metric.

\begin{figure}[htbp]
  \centering
  \includegraphics[width=0.49\textwidth]{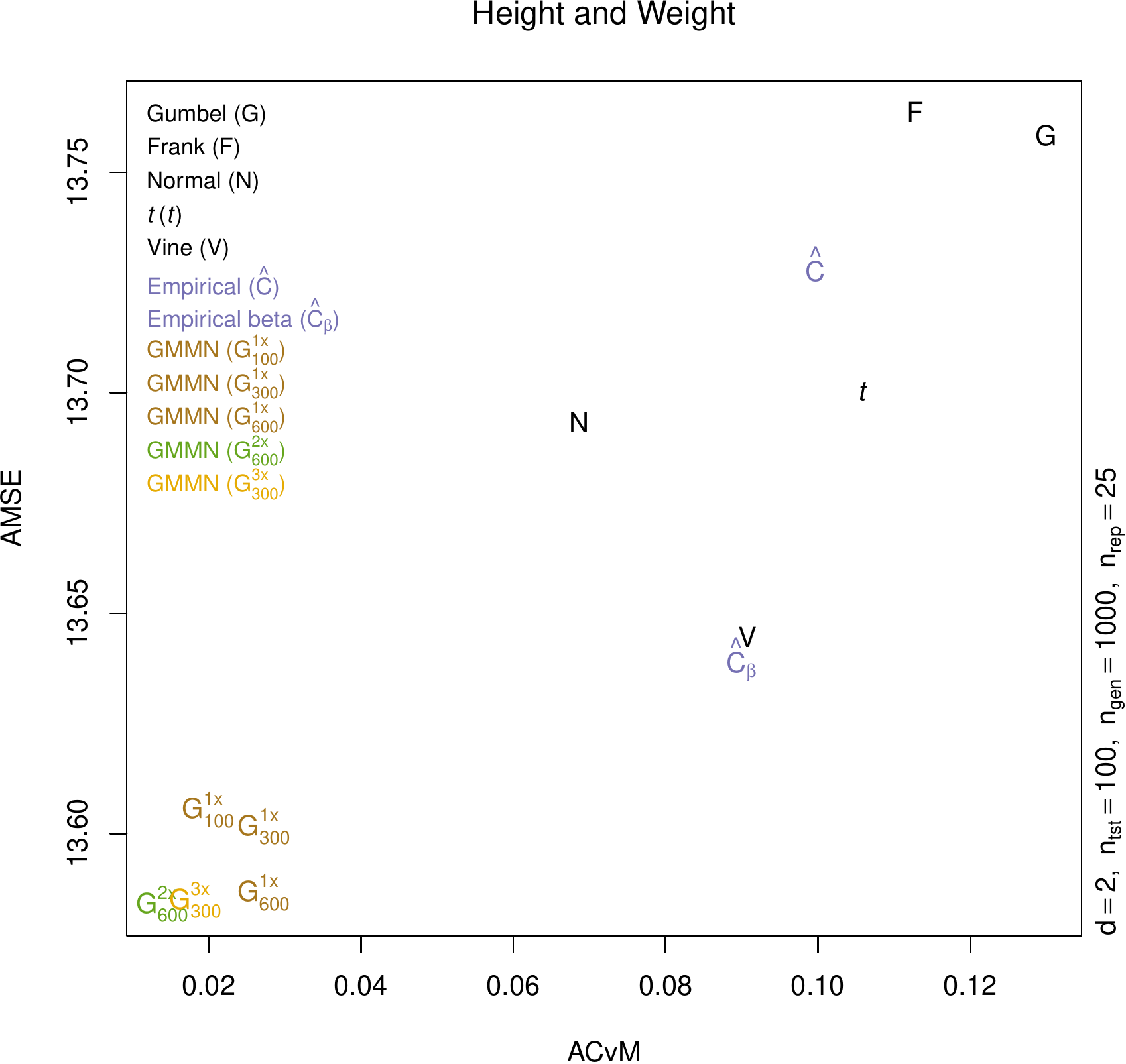}
  \hfill
  \includegraphics[width=0.49\textwidth]{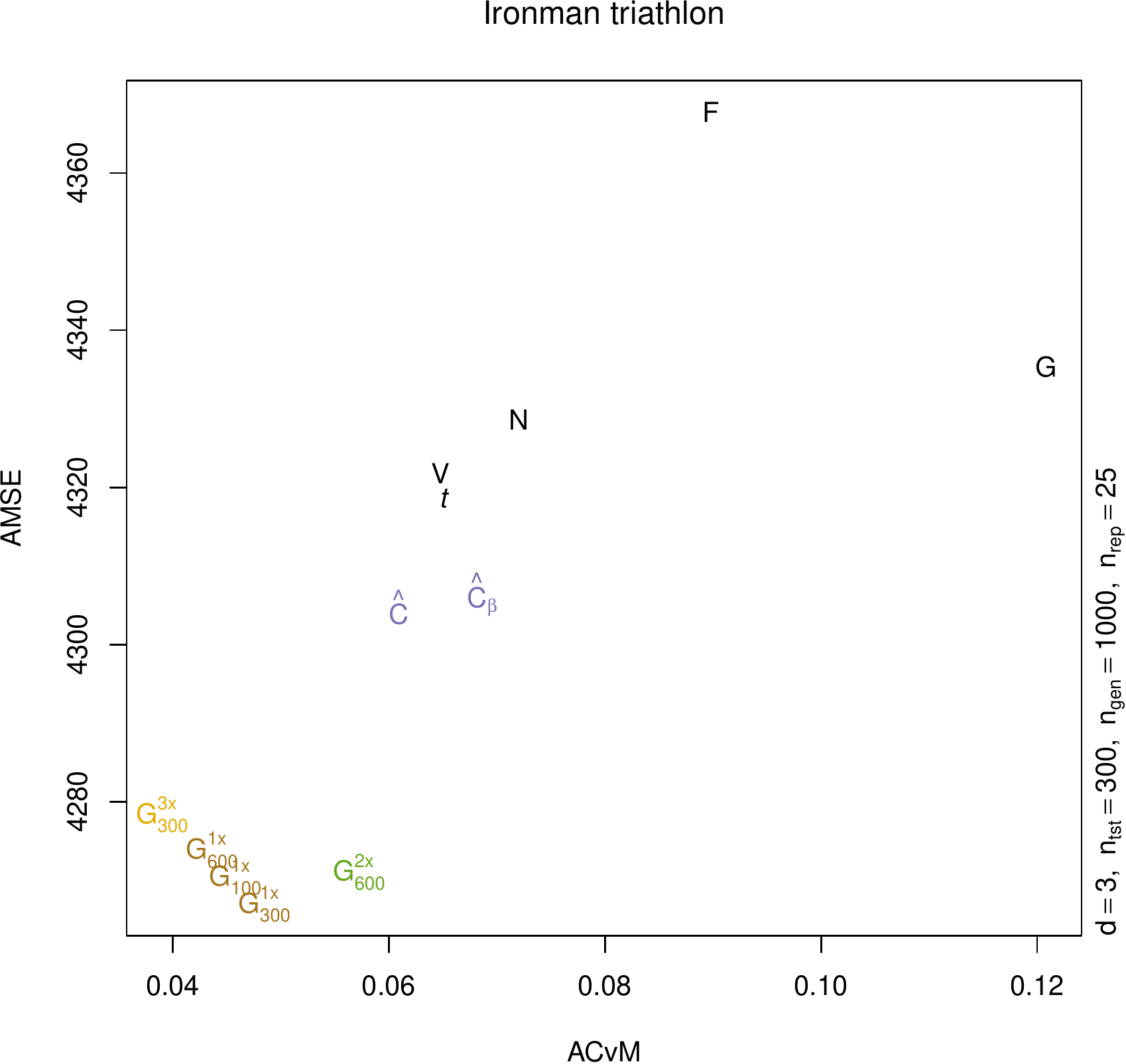}\\[3mm]
  \includegraphics[width=0.49\textwidth]{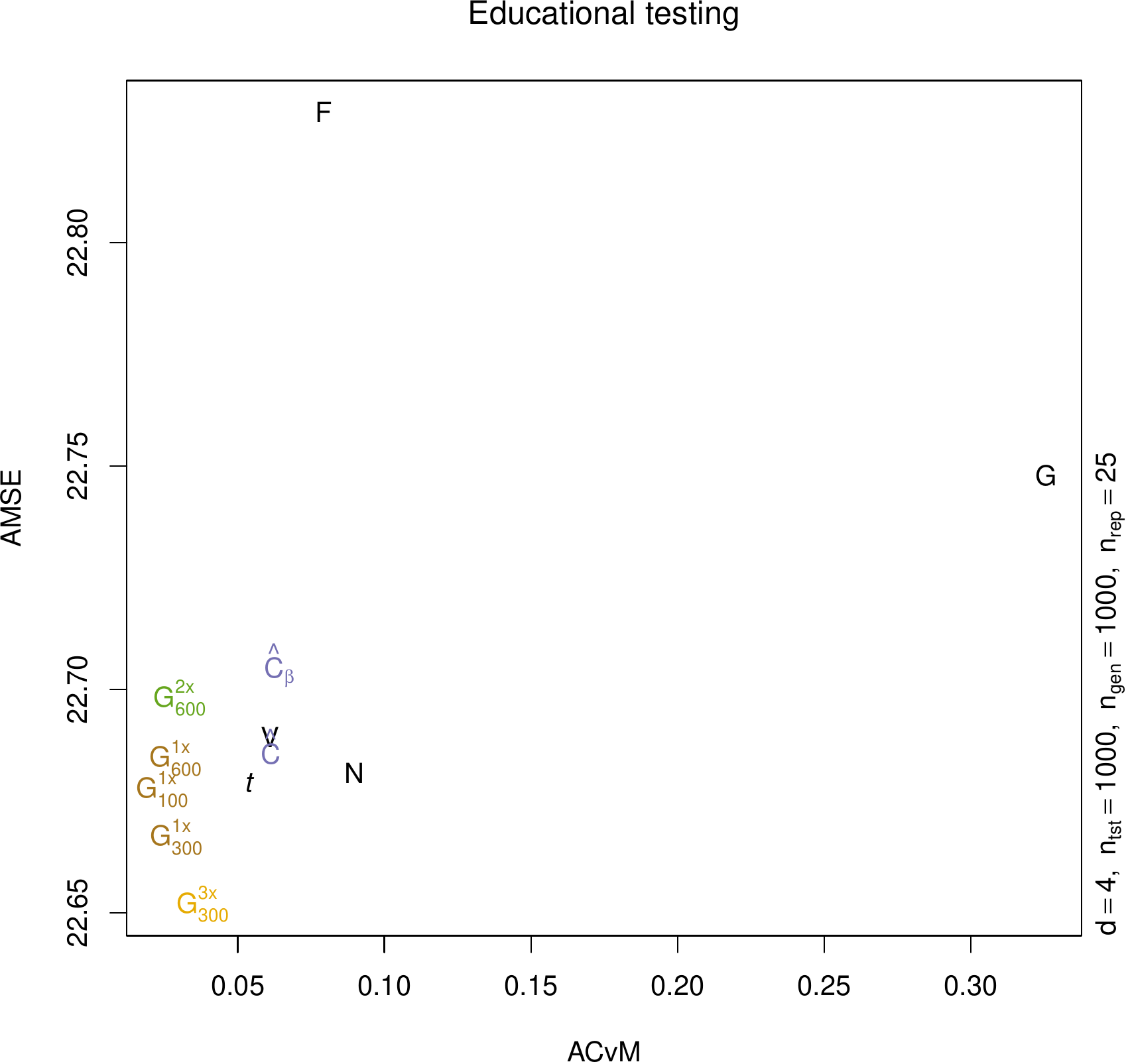}
  \hfill
  \includegraphics[width=0.49\textwidth]{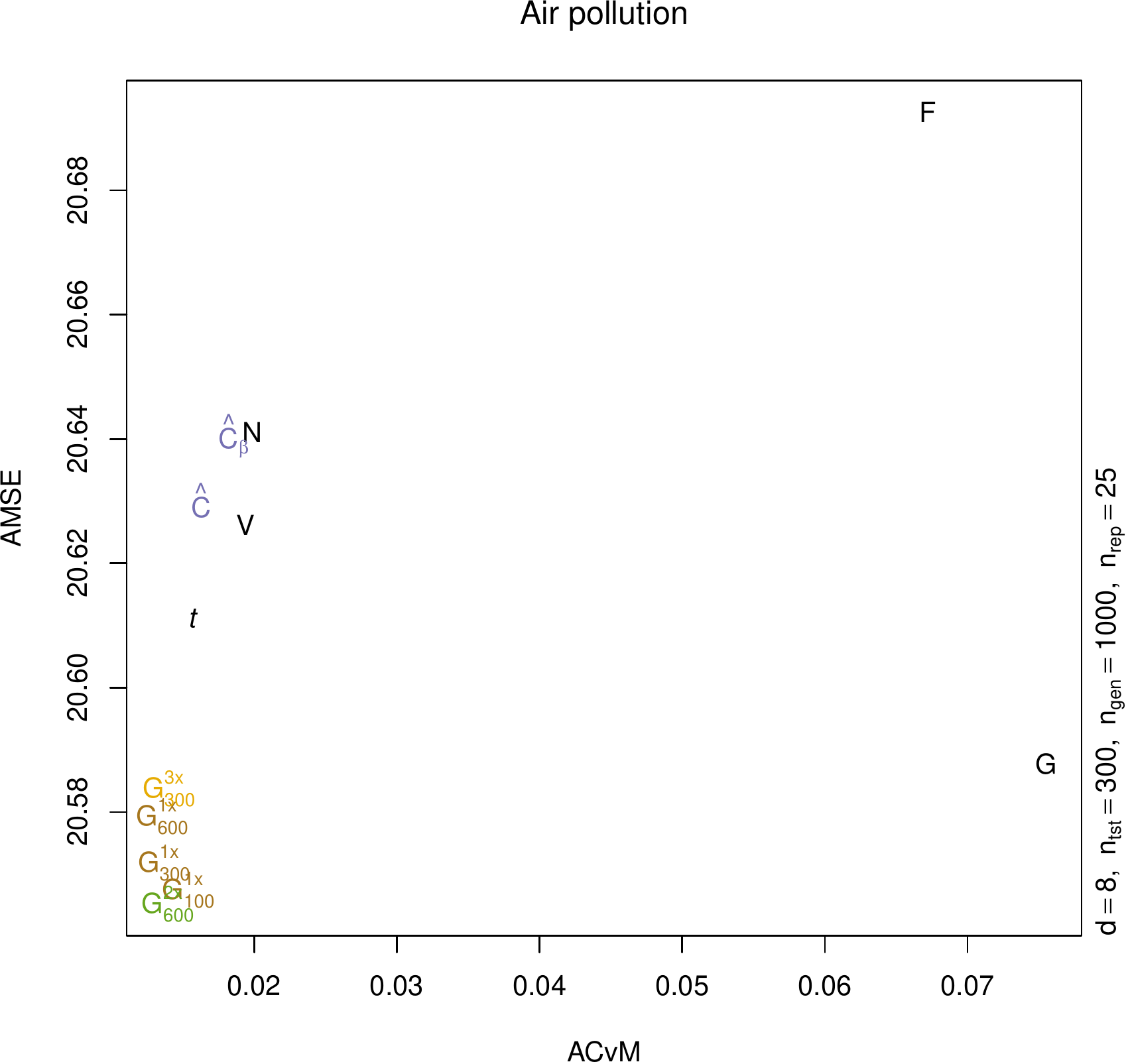}
  \caption{Model assessments for the height and weight data (top left), the
    ironman triathlon data (top right), the educational data (bottom left) and
    the air pollution data (bottom right). The $\text{ACvM}$ is evaluated with
    $\nrep=25$ replications, and the $\AMSE$ is evaluated based on $\ngen=1000$
    samples.}\label{fig:examples}
\end{figure}

Figure~\ref{fig:examples} shows scatter plots of AMSE versus ACvM for the four
examples considered.
From these plots, we observe that, in most cases, the samples generated from the five neural
network models more closely match the underlying dependence $C_{\ntst}$ of the
test data than those generated from competing copula models. Moreover, this
better dependence modeling (as assessed by ACvM) does, in most cases, translate
into better probabilistic predictions (as assessed by the AMSE metric). The five
RafterNets therefore typically produce better empirical predictive distributions
when compared with various RafterCops.

\section{Conclusion and outlook}
We suggested a fully nonparametric approach, the RafterNet, for making
probabilistic predictions in multi-response regression problems. First, random
forests are used to model the mean of each response variable $\E(X_j)$ as
flexible functions of the covariates. Then, empirical distributions are used to
model the marginal distributions of the residuals $X_j-\E(X_j)$. Finally, as a
novel contribution, we introduced generative neural networks to model the
joint distribution of $(F_1(X_1-\E(X_1)),\dots,F_d(X_d-\E(X_d)))$ in place of
conventional copulas. The flexibility of RafterNets were showcased in four
different data examples, where we demonstrated how using neural networks yielded superior
probabilistic predictions compared to various copula models.

It would be desirable to relax the restriction in
Equation~\eqref{eq:specialSklar} that $C$ is generic for all $k$ and does not
depend on the covariate. However, $C$ is much harder to model nonparametrically
than $\theta_j(\cdot)$ and $F_j$.  If, for a random variable $\bm{U}\in\IR^d$ (here,
with uniform margins), its distribution $C_{\bm{z}}(u_1,\dots,u_d)$ behaves
differently over the space of covariates $\bm{z}$, then, in order to understand
the difference between $C_{\bm{z}}$ and $C_{\bm{z}'}$ without any parametric
assumption, it will be necessary to have observed $\bm{U}$ near both $\bm{z}$
and $\bm{z}'$ a relatively large number of times. If not, then surely some sort
of smoothness assumptions will be required to describe the behavior of
$C_{\bm{z}}$ in the space of $\bm{z}$. This is a challenging problem in the
realm of generative neural networks that is open for future research.

\Urlmuskip=0mu plus 1mu\relax%
\printbibliography[heading=bibintoc]

\appendix
\section{The air pollution data with different marginal models}
\label{appdx:marg}
In Section~\ref{sec:examples}, we have naturally focused on empirical
results produced by performing
\eqref{eq:deMargin_nonparam}--\eqref{eq:addMargin_nonparam}. In this appendix,
we use the air pollution dataset to illustrate the effect of performing
\eqref{eq:deMargin_param}--\eqref{eq:addMargin_param} instead. Specifically,
for every $j=1,\dots,d$, we model the marginal distribution
$F_j(x_{k,j};\theta_j(\bm{z}_k))$ as a gamma distribution having density function
\begin{align}
  f_j(x_{k,j};\theta_j(\bm{z}_k)) = \frac{\theta_j(\bm{z}_k)^{\alpha_j}}{\Gamma(\alpha_j)}
  x_{k,j}^{\alpha_j-1} e^{-\{\theta_j(\bm{z}_k) x_{k,j}\}}
  \quad\text{with}\quad \log\theta_j(\bm{z}_k) = \bm{\beta}_j\T\bm{z}_k,
  \label{eq:gammaGLM}
\end{align}
or, in short, as $\Gamma(\alpha_j,\exp(\bm{\beta}_j\T\bm{z}_k))$.
But as in Section~\ref{sec:examples}, we still focus mainly on comparing our
approach of using neural networks to model the copula $C$ (referred to as
``GLMNets'' in accordance with ``RafterNets'') with the conventional approach
of using parametric copulas (referred to as ``GLMCops'' in accordance with
``RafterCops'').

\begin{figure}[htbp]
  \centering
  \includegraphics[width=0.49\textwidth]{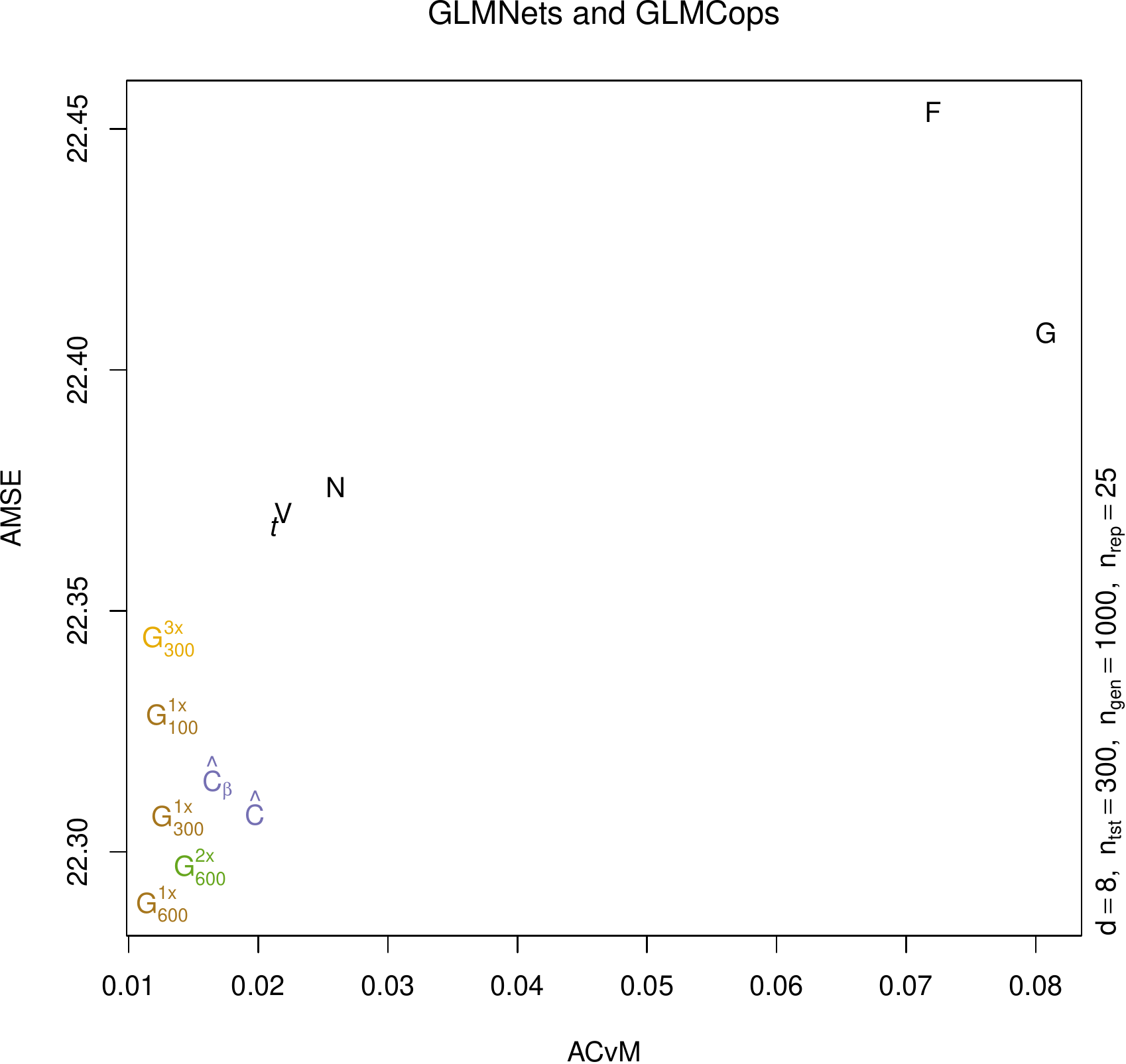}%
  \hfill
  \includegraphics[width=0.49\textwidth]{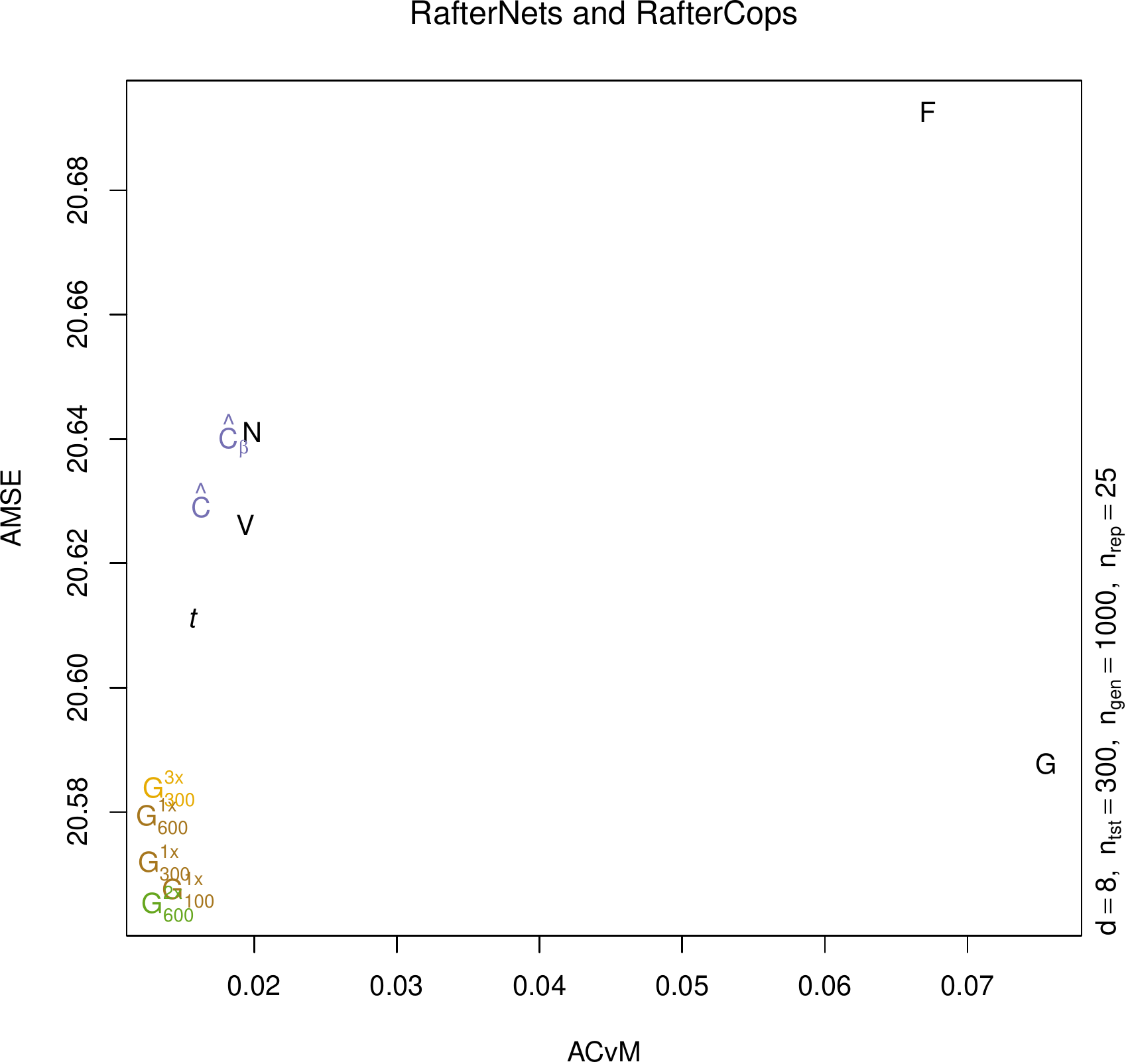}%
  \caption{Model assessments of GLMNets and GLMCops (left) and RafterNets
    and RafterCops (right) for air pollution data. The $\text{ACvM}$ is
    evaluated with $\nrep=25$ replications, and the $\AMSE$ is evaluated
    based on $\ngen=1000$ samples.}\label{fig:examples:RFvsGLM}
\end{figure}

The left panel of Figure~\ref{fig:examples:RFvsGLM} shows that GLMNets (using
different neural network architectures) generally outperform GLMCops (using
different parametric copulas) in both metrics, so there is clearly benefit in
using neural nets rather than parametric copulas to model dependence
regardless of how the marginal distributions are modeled. This is the main
point of our article.

The right panel of Figure~\ref{fig:examples:RFvsGLM} merely reproduces the
bottom-right panel of Figure~\ref{fig:examples}, for ease of direct
comparison. After comparing the y-axis with the left panel, we can see that
RafterNets and RafterCops outperform GLMNets and GLMCops in the AMSE
metric. This is not surprising since random forests are more flexible than GLMs,
but there are exceptions, for example, if the true marginal distributions are
very close to \eqref{eq:gammaGLM}, then one would
expect GLMNets and GLMCops to be superior, but that is both obvious and not
the main point of our article.

\section{Evidence that VAEs do not properly learn dependence}
\label{appdx:vae}
We now provide some empirical evidence to support what we have said in
Section~\ref{sec:vae}, namely that VAEs are not effective for our specific
learning task. We consider the simple task of learning to generate
from a few well-known copula models. Specifically, we learn to generate from
$2$-dimensional and $10$-dimensional Clayton and $t_4$ copulas with pairwise
Kendall's tau set to $\tau=0.5$, using training samples of size $\ntrn=50\,000$
from the true copula.

First, we use the same five neural networks from Section~\ref{sec:examples},
that is $G^{\ell\text{x}}_h$ with $\{\ell,h\}= \{1,100\}$, $\{1,300\}$,
$\{1,600\}$, $\{2,600\}$, and $\{3,300\}$. Then, we perform the
same task with VAEs. We use the same five architectures, for both the
encoder network (mapping from the training sample to $\bm{V}$) and the decoder
network (mapping from $\bm{V}$ to the output) within each VAE. For VAEs, we also
experiment with different dimensions of $\bm{V}$. We denote each VAE
architecture by $V^{\ell\text{x}}_{h,d'}$, where $d'$ is the dimension of
$\bm{V}$.

\begin{figure}[htbp]
  \centering
  \includegraphics[width=\textwidth]{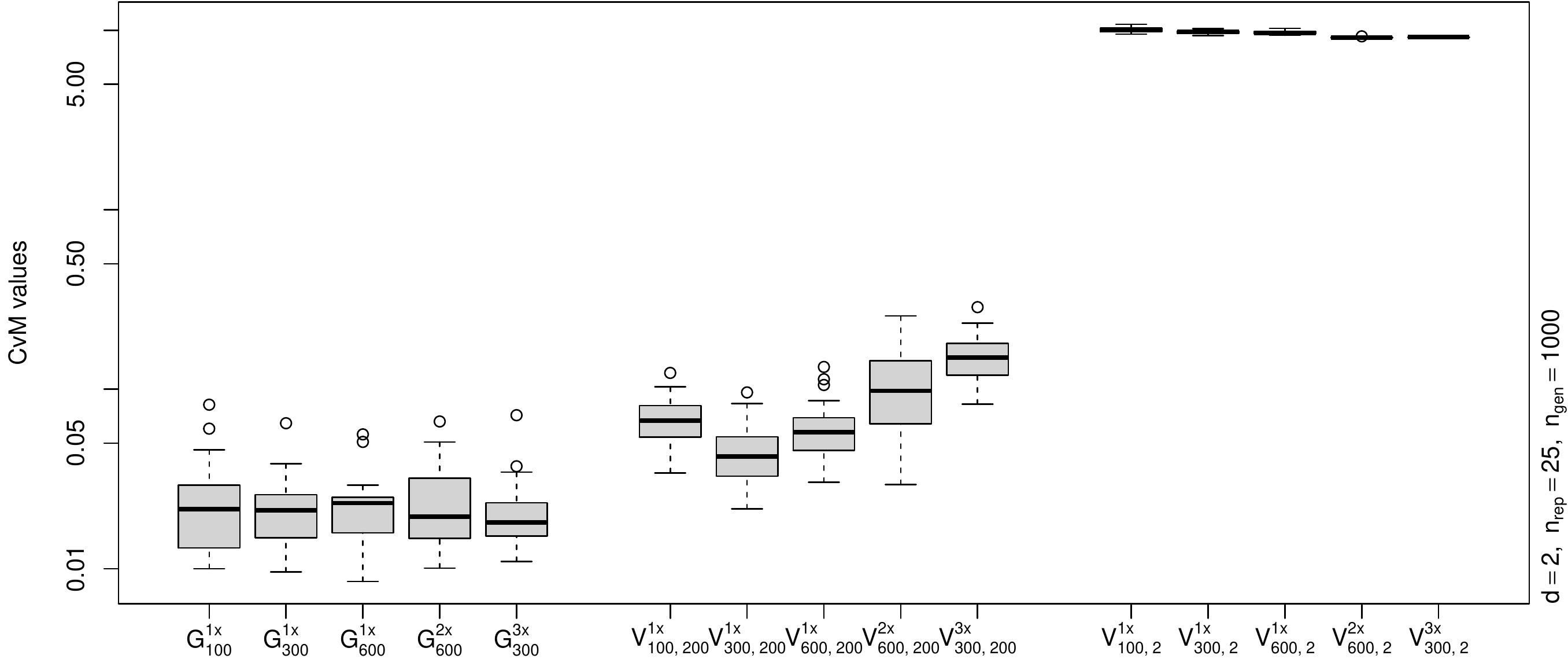}
  \\[4mm]
  \includegraphics[width=\textwidth]{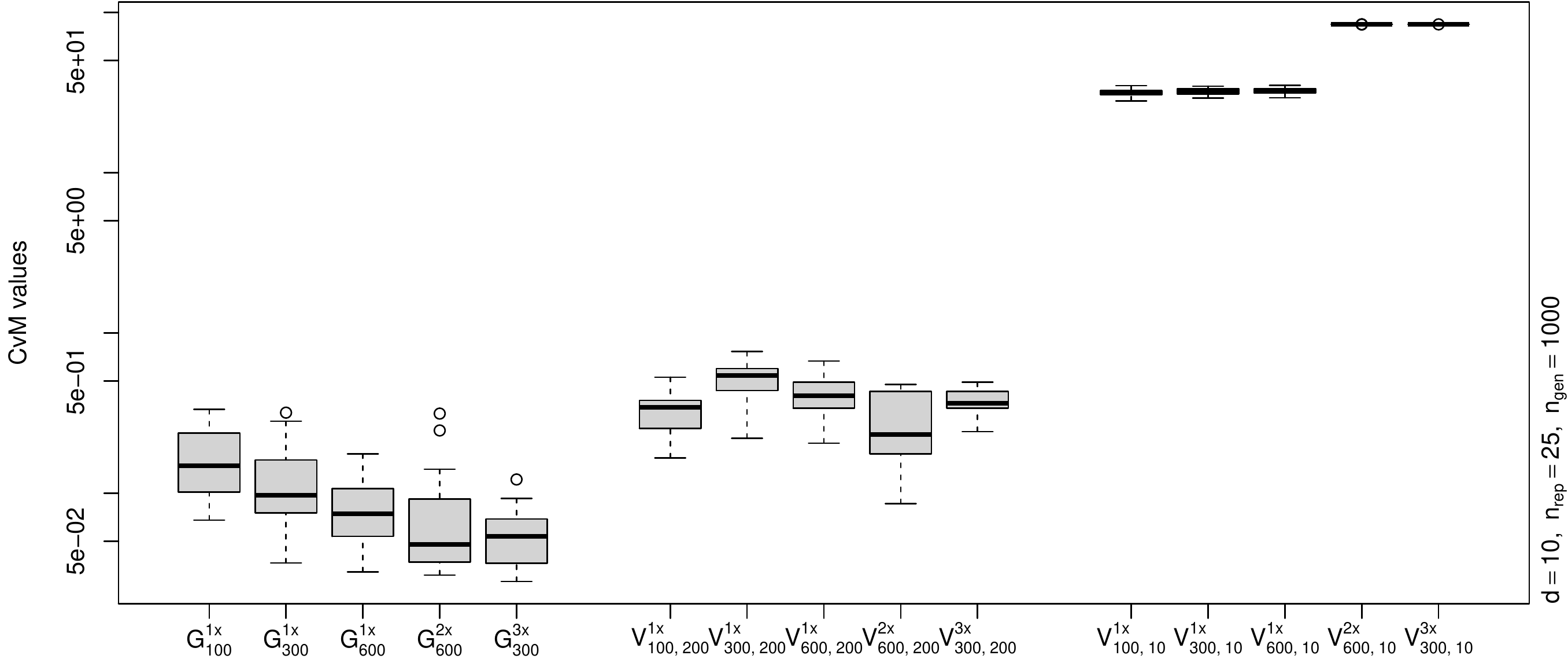}
  \caption{Boxplot of CvM values evaluating samples generated by our five neural
    network models and the ten considered VAE models, which were trained on
    $d=2$-dimensional (top) and $d=10$-dimensional (bottom row) Clayton copulas
    with $\tau=0.5$.}\label{fig:GNN:comp:Clayton}
\end{figure}

\begin{figure}[htbp]
  \centering
  \includegraphics[width=\textwidth]{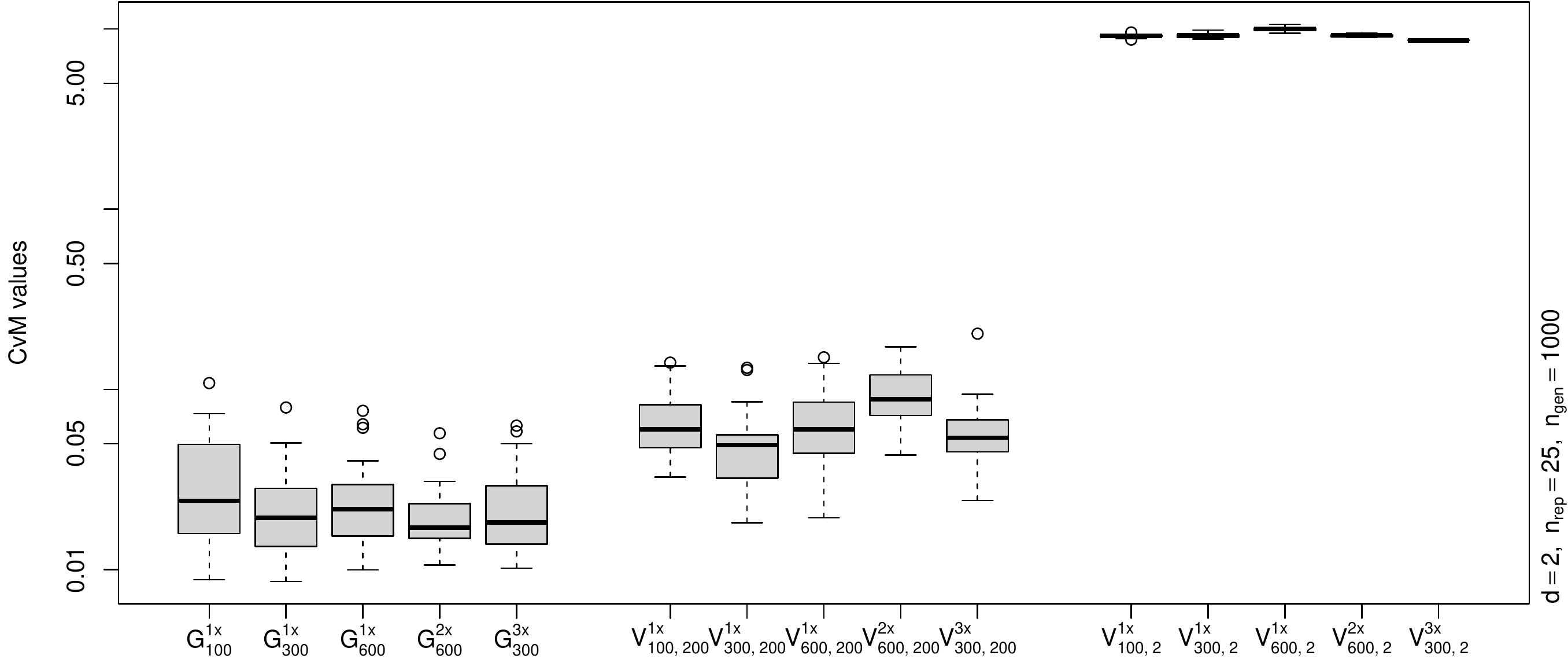}
  \\[4mm]
  \includegraphics[width=\textwidth]{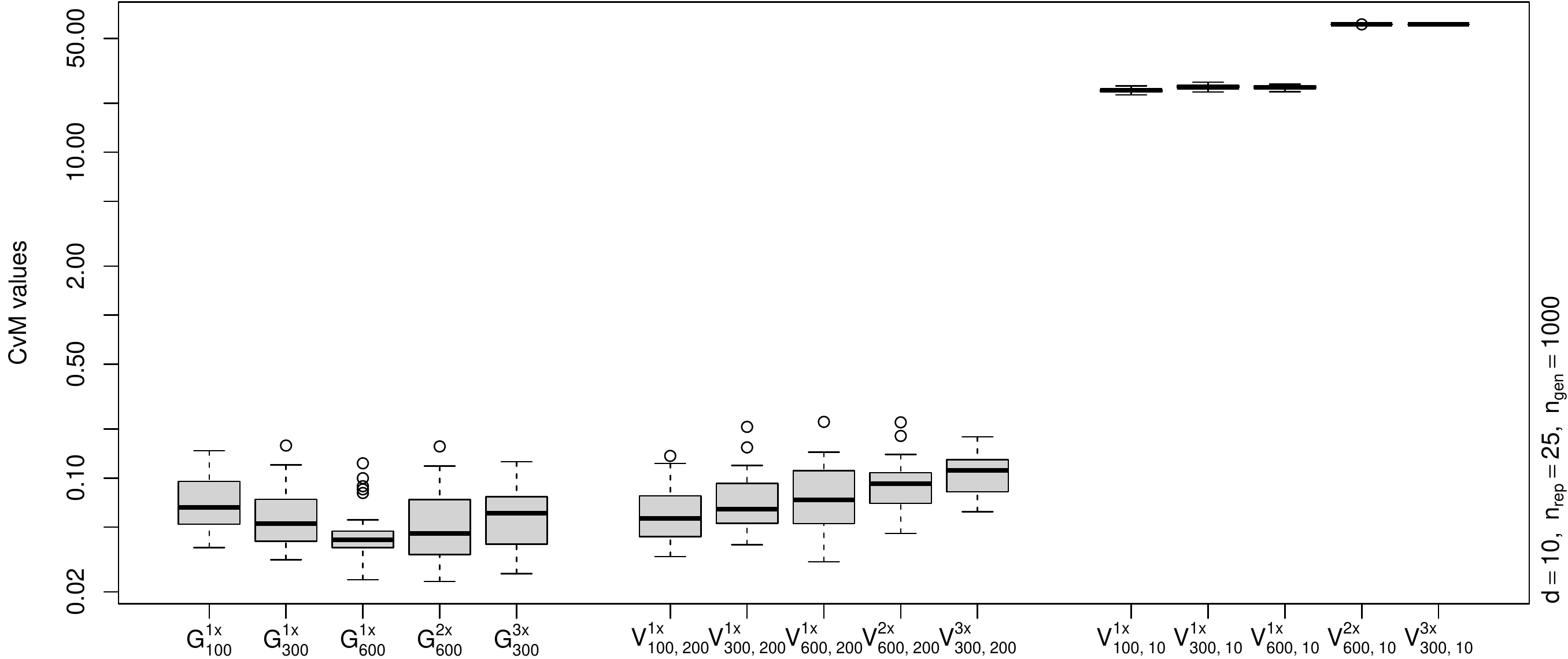}
  \caption{Boxplot of CvM values evaluating samples generated by our five neural
    network models and the ten considered VAE models, which were trained on
    $d=2$-dimensional (top) and $d=10$-dimensional (bottom) $t_4$ copulas with
    $\tau=0.5$.}\label{fig:GNN:comp:t}
\end{figure}

After each $G^{\ell\text{x}}_h$ and $V^{\ell\text{x}}_{h,d'}$ is trained, each
for $\nepo=300$ epochs which we are empirically sure is more than necessary for
a fair comparison, we use it to generate $B=25$ samples, each of size
$\ngen=1000$. We then assess the quality of \emph{each} generated sample against the
ground truth using the one-sample Cram\'er--von Mises statistic, defined as
\begin{align*}
  \CvM = \int_{[0,1]^{d}} \ngen(C_{\ngen}(\bm{u})-C(\bm{u}))^2\,\rd C_{\ngen}(\bm{u}), %
\end{align*}
where $C_{\ngen}$ is the empirical copula of the $\ngen$ generated samples and
$C$, the true copula we are trying to learn. These 25 measurements of
$\CvM$ are summarized by boxplots in Figure~\ref{fig:GNN:comp:Clayton} for
learning Clayton copulas and in Figure~\ref{fig:GNN:comp:t} for learning $t_4$
copulas.

When $d'=d$, the samples generated by VAEs are significantly worse than those
generated by the neural networks we have adopted (which also use $d'=d$). Even
when $d'\gg d$, VAEs still cannot learn the distributions properly, although
their performance improves. However, it should be immediately clear to anyone
that, even if the VAEs could be made to perform equally well as our neural
networks, perhaps by using a very large $d'$ indeed, it would still be highly
inefficient if, in order to learn
a $2$-dimensional distribution, one must first ``embed'' it onto a manifold that
lies in $\IR^{200}$ or even $\IR^{500}$.

\section{Demonstration in \R\ with height and weight data}\label{sec:R:implementation}
In this section, we illustrate how to implement our RafterNet approach in \R\
based on the height and weight dataset. To begin with, we load the required \R\
packages; note that your system needs Python's Keras and TensorFlow installed
for neural network training and evaluation.
\begin{Sinput}
library(randomForest) # for regression modeling
library(copula) # (only) for pseudo-observations and their transformations
library(keras) # interface to Keras Python package (high-level neural network API)
library(tensorflow) # interface to TensorFlow Python package
library(gnn) # for generative neural network modeling
\end{Sinput}
As mentioned in Section~\ref{sec:datasets}, the dataset can be found on the
webpage \cite{mcelreath2020} under the name \texttt{Howell1.csv}. The following
chunk downloads and reads this dataset.
\begin{Sinput}
## Loading height and weight data
## (see https://github.com/rmcelreath/rethinking/tree/master/data -> Howell1.csv)
url <- "https://raw.githubusercontent.com/rmcelreath/rethinking/master/data/"
file <- "Howell1.csv"
if(!file.exists(file)) # if not yet existing, download the dataset
    download.file(paste0(url, file), destfile = file)
raw <- read.csv("Howell1.csv", sep = ";") # read the data frame
dat <- as.matrix(raw) # convert raw to a numeric matrix
\end{Sinput}
To model the distribution of the height and
weight of individuals conditional on their age and sex we use observations
from $\ntrn=444$ individuals to train the RafterNet and the remaining
$\ntst=100$ observations as a test sample.
\begin{Sinput}
## Split the dataset into training and test data
n.tst <- 100 # number of test samples
set.seed(271) # for reproducibility
tst.obs <- sample(1:nrow(dat), size = n.tst) # indices of observations in test data
dat.trn <- dat[-tst.obs,] # training data
dat.tst <- dat[ tst.obs,] # test data
## Convenient variable definitions
X.trn <- dat.trn[, c("height", "weight")] # response variables in training data
z.trn <- dat.trn[, c("age", "male")] # covariates in training data
X.tst <- dat.tst[, c("height", "weight")] # response variables in test data
z.tst <- dat.tst[, c("age", "male")] # covariates in test data
\end{Sinput}
First, we separately model the mean height and mean weight $\E(X_{k,j})$,
$j=1,2$, of individuals as flexible functions of the covariates $\bm{z}_k$,
$k=1,\dots,\ntrn$, (age and sex of individuals) using random forests
$\theta_j(\bm{z}_k)$.
\begin{Sinput}
## Fit a random forest for each response variable
d <- ncol(X.trn) # dimension
raft.fits <- lapply(1:d, function(j) randomForest(y = X.trn[,j], x = z.trn))
\end{Sinput}
Next, we use the fitted random forests $\widehat{\theta}_j(\bm{z})$, $j=1,2$, to
compute the realized residuals
$\widehat{R}_{k,j} = X_{k,j} - \widehat{\theta}_j(\bm{z}_k)$, $k=1,\dots,\ntrn$,
$j=1,2$. Thereafter we nonparametrically model the marginal distributions of the
realized residuals $\widehat{R}_{k,j}$, $k=1,\dots,\ntrn$, $j=1,2$, by
computing the pseudo-observations $\widehat{\bm{U}}_k$, $k=1,\dots,\ntrn$, as
described in \eqref{eq:deMargin_nonparam}.
\begin{Sinput}
## Compute realized residuals and their pseudo-observations
R.hat <- sapply(1:d, function(j) X.trn[,j] - raft.fits[[j]]$predicted) # residuals
U.hat <- pobs(R.hat) # compute the corresponding pseudo-observations
\end{Sinput}

Now we can model the pseudo-observations
$\widehat{\bm{U}}_{k}$, $k=1,\dots,\ntrn$, using a neural network
$G$. For this illustration, we work with a neural network with a single hidden
layer consisting of 100 neurons. Due to its non-expensive and non-vanishing
gradients, we use a ReLU activation function
$\phi(x)=\max\{0,x\}$ in the hidden layer. Since our target output (the
pseudo-observations) lie in $(0,1)^2$, we use a sigmoid activation function
$\phi(x)=1/(1+e^{-x})$ in the output layer. This neural network architecture
can be specified as follows.
\begin{Sinput}
## NN setup
n.trn <- nrow(U.hat) # number of training observations
dim.in.out <- d # dimension of the input and output layers of the NN
dim.hid <- 100 # dimension of the (single) hidden layer
NN.dim <- c(dim.in.out, dim.hid, dim.in.out) # NN architecture
## Define the NN model
NN.model <- FNN(dim = NN.dim, # dimension of NN layers
                activation = c("relu", "sigmoid"), # activation functions
                batch.norm = TRUE, # adding batch normalization layer
                dropout.rate = 0.1, # dropout rate for dropout layer(s)
                loss.fun = "MMD") # loss function used
\end{Sinput}
We take as input to the neural network, a sample
$\{\bm{V}_l\}_{l=1}^{\ntrn}$ from
$\N(\bm{0},I_{2})$. As explained in Section~\ref{sec:gnnTraining}, we then train
the neural network based on the optimization problem described in
\eqref{eq:opt_K}, where the kernel function
$K$ is a mixture of Gaussian kernels with different bandwidth parameters
$\bm{h}=(0.001, 0.01, 0.15, 0.25, 0.50,
0.75)$. Additionally, we use batch normalization and dropout regularization
(with dropout rate
$0.1$) in the hidden layer to help control possible overfitting while
training. The neural network is trained for 1000 epochs and the network with the
best weights over the entire training process is selected.
\begin{Sinput}
## Define training hyperparameters
n.epo <- 1000 # number of epochs used for NN training
n.bat <- n.trn # batch size used for NN training
## Training of the NN
NN <- fitGNNonce(NN.model, # model to be trained
                 data = U.hat, # training data
                 batch.size = n.bat, # batch size
                 n.epoch = n.epo, # number of epochs
                 prior = rPrior(n.trn, copula = indepCopula(dim.in.out), # prior
                                qmargins = qnorm), # N(0,1) margins of prior
                 file = "GNN_height_weight.rda",
                 ## The following argument specifies to take the best weights
                 ## over n.epo-many epochs
                 callbacks = callback_early_stopping(monitor = "loss",
                                                     min_delta = 0,
                                                     patience = n.epo,
                                                     restore_best_weights = TRUE))
\end{Sinput}

\begin{figure}[htbp]
  \centering
  \includegraphics[width=0.49\textwidth]{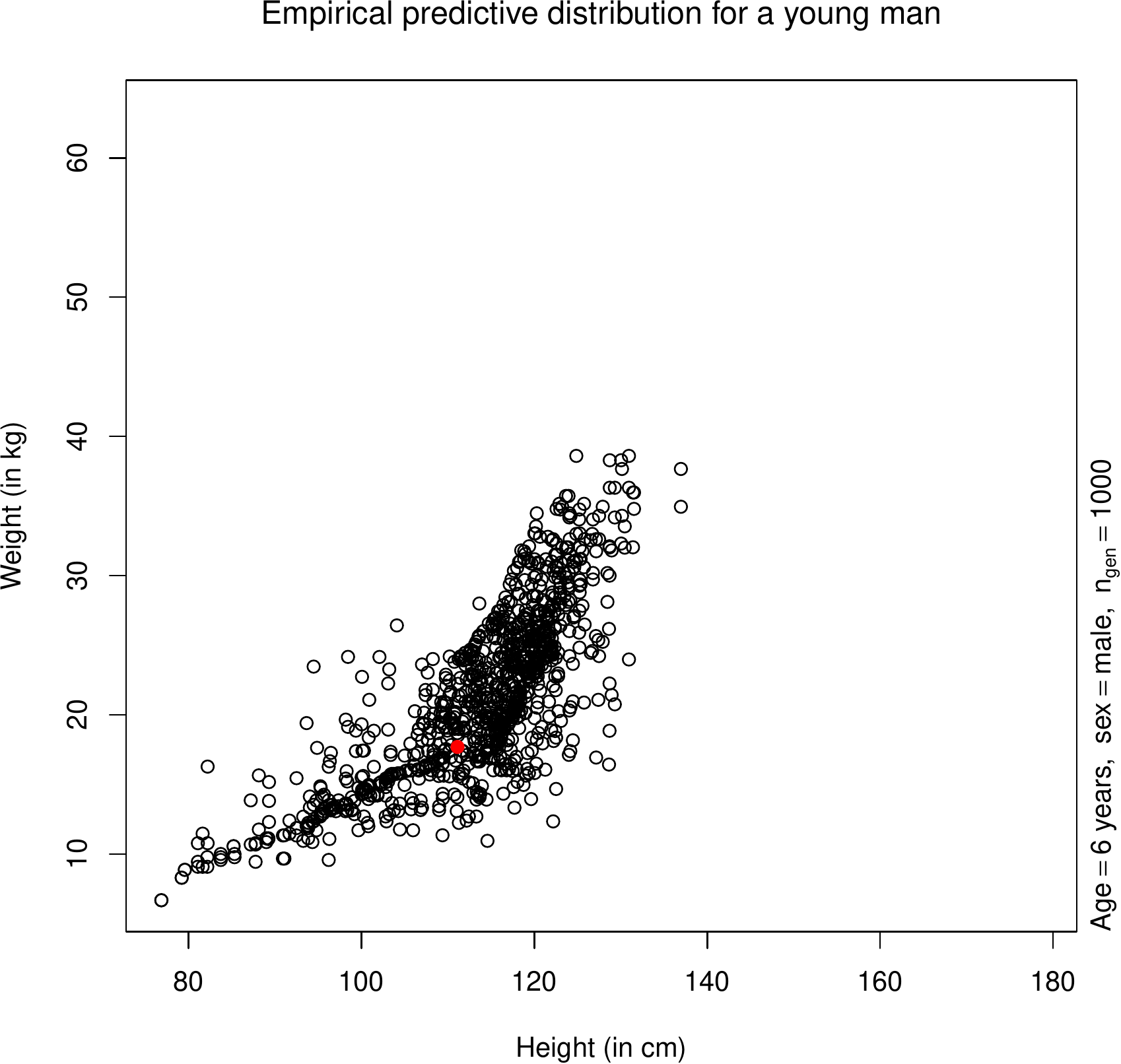}
  \hfill
  \includegraphics[width=0.49\textwidth]{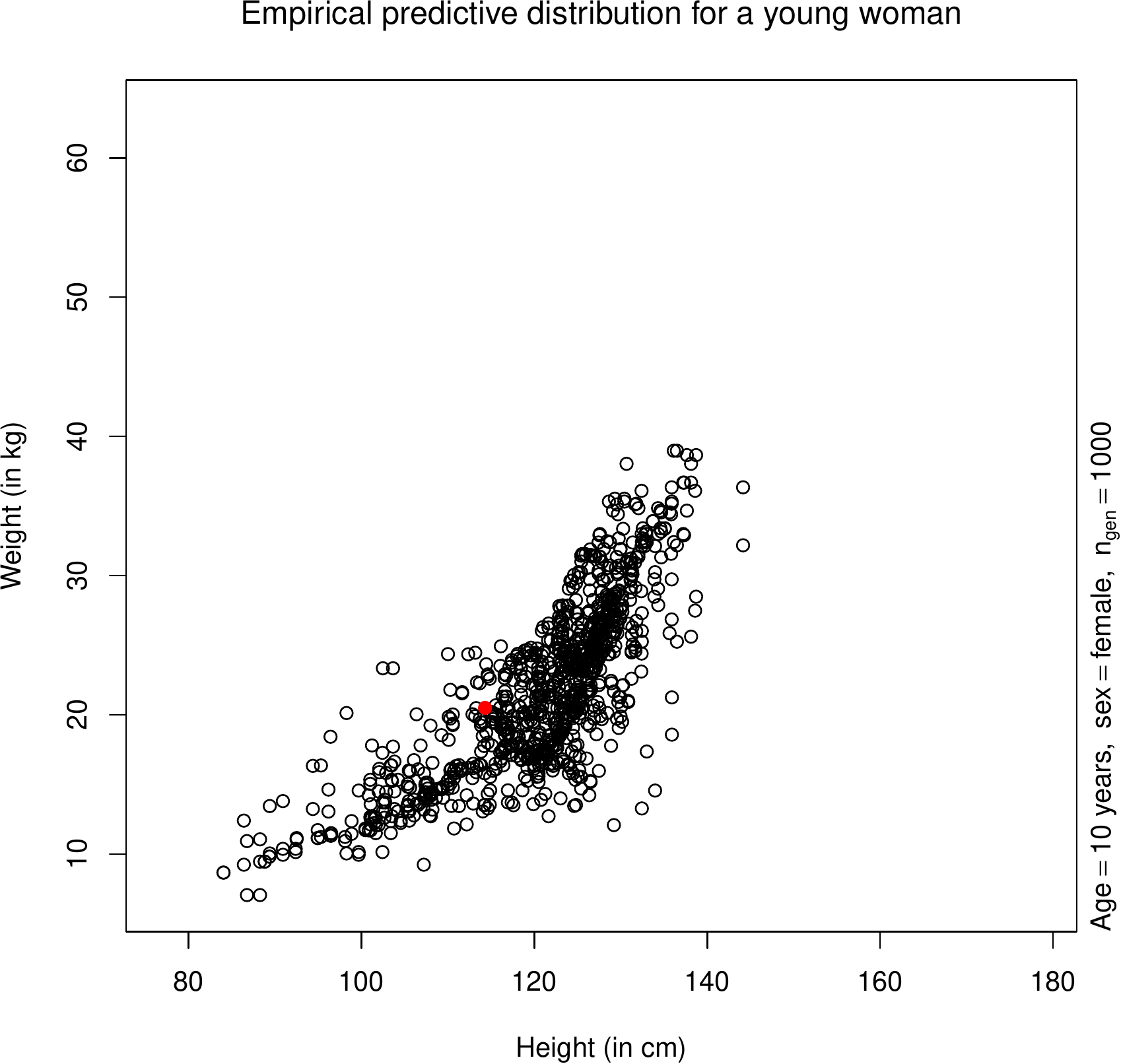}\\[3mm]
  \includegraphics[width=0.49\textwidth]{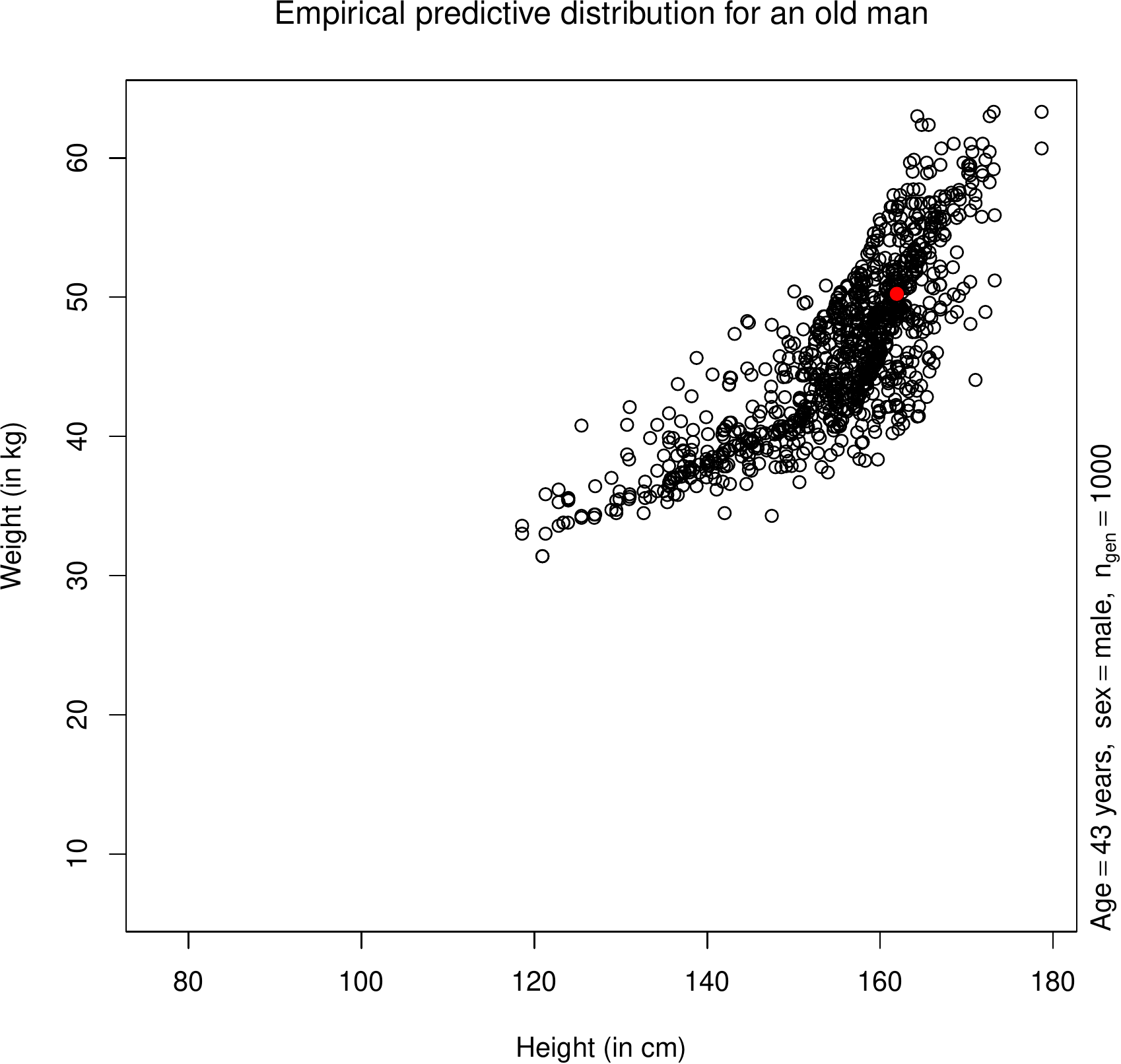}
  \hfill
  \includegraphics[width=0.49\textwidth]{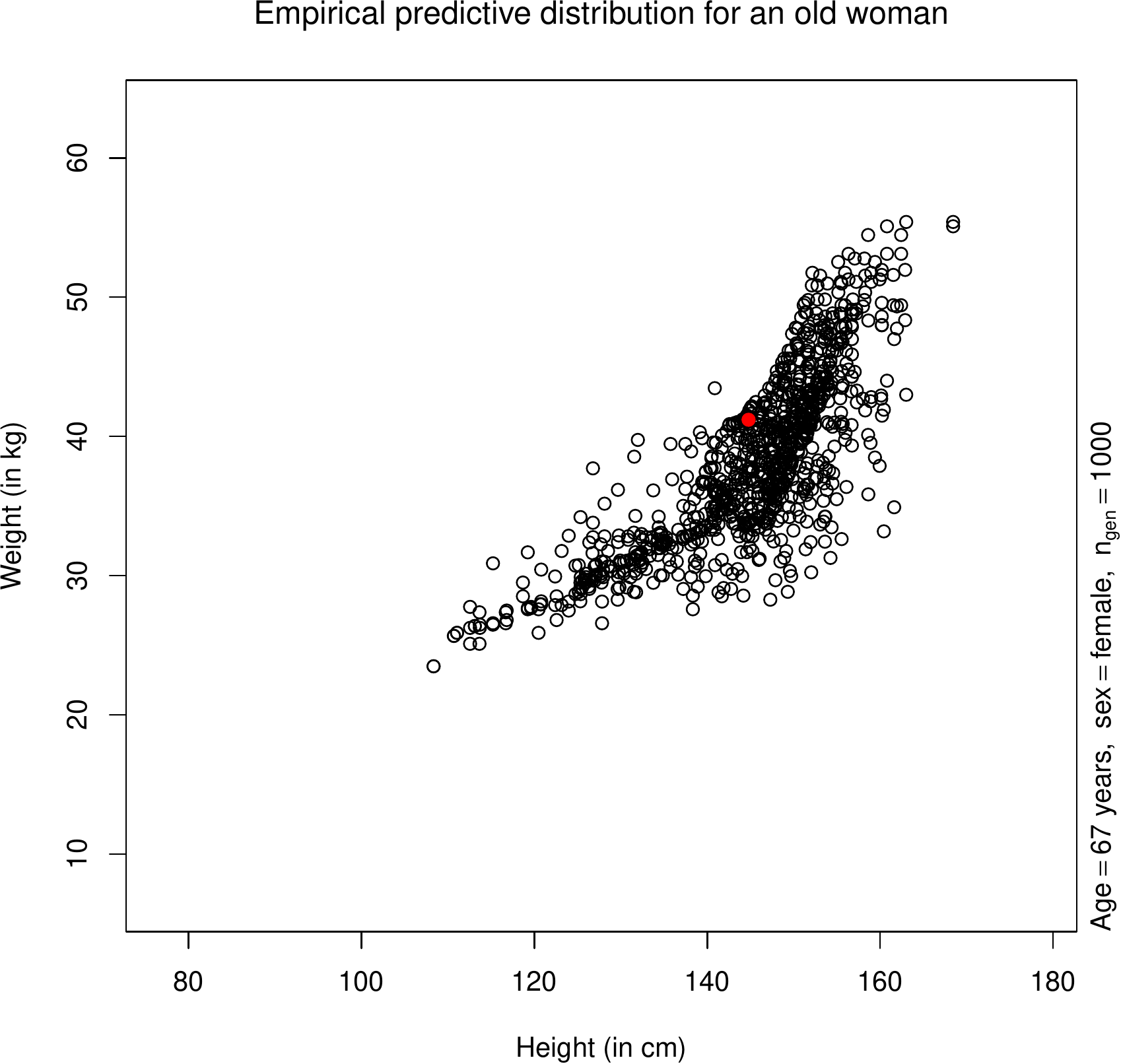}
  \caption{Scatter plots of empirical predictive distributions produced by
    RafterNet, one for each of four given covariate realizations along with
    the corresponding true response realizations (red points) for the height and
    weight dataset.}\label{fig:htwt:four:cases}
\end{figure}

We can now use our trained RafterNet, to make probabilistic predictions for a
given covariate $\bm{z}_k$ by following the procedure summarized in
Algorithm~\ref{tab:rafternet_predict}. For our illustration here, we select the
observations corresponding to six year old males from our test sample (there was
one such observation) and make a probabilistic prediction using $\ngen=1000$
samples of the height and weight for such individual(s).
\begin{Sinput}
## Generate samples from the trained NN
n.gen <- 1000 # number of samples to generate
U.gen <- rGNN(NN, size = n.gen, pobs = TRUE) # new observations from hat(C)
## Convert to empirical margins
R.pred <- toEmpMargins(U.gen, x = R.hat) # corresponding residuals
## Probabilistic prediction using RafterNet for the covariate (six year old male)
z1 <- c(age = 6.0, male = 1) # covariate combination six-year-old male
stopifnot(sum( # check that there is exactly one such covariate combination
    apply(z.tst, 1, function(x) all(x == z1))) == 1)
X.pred1 <- sapply(1:d, # transform to the right margins
                  function(j) predict(raft.fits[[j]], newdata = z1) + R.pred[,j])
\end{Sinput}
We repeat the prediction step described in the code above for three additional
sets of observations from the test sample, corresponding to 10-year-old females,
43-year-old males, and 67-year-old females,
respectively.
\begin{Sinput}
## Probabilistic predictions for other covariate combinations (checks also passed)
R.pred2 <- toEmpMargins(rGNN(NN, size = n.gen, pobs = TRUE), x = R.hat)
z2 <- c(age = 10, male = 0)
X.pred2 <- sapply(1:d, function(j)
    predict(raft.fits[[j]], newdata = z2) + R.pred2[,j])
R.pred3 <- toEmpMargins(rGNN(NN, size = n.gen, pobs = TRUE), x = R.hat)
z3 <- c(age = 43, male = 1)
X.pred3 <- sapply(1:d, function(j)
    predict(raft.fits[[j]], newdata = z3) + R.pred3[,j])
R.pred4 <- toEmpMargins(rGNN(NN, size = n.gen, pobs = TRUE), x = R.hat)
z4 <- c(age = 67, male = 0)
X.pred4 <- sapply(1:d, function(j)
    predict(raft.fits[[j]], newdata = z4) + R.pred4[,j])
\end{Sinput}
Figure~\ref{fig:htwt:four:cases} displays the probabilistic predictions of the
heights and weights for each of these four sets of covariates along with the
corresponding true heights and weights of such individuals (represented by red
points). From these plots we observe that the empirical predictive distributions
indeed roughly concentrate around each of the corresponding height and weight
observations.

For each of these four individuals defined by the covariate $\bm{z}_k = (\text{age}_k, \text{sex}_k)$,
  we can also predict joint tail probabilities such as $\P(X_{k,1}=\text{height}_k>c_1, X_{k,2}=\text{weight}_k<c_2)$
  for any fixed constants $(c_1, c_2)$. Here are some examples.
\begin{Sinput}
## Predict joint probabilities
mean(X.pred1[,1] > 116 & X.pred1[,2] < 21) # ~= 0.1
mean(X.pred2[,1] > 116 & X.pred2[,2] < 21) # ~= 0.247
mean(X.pred3[,1] < 158 & X.pred3[,2] > 46) # ~= 0.121
mean(X.pred4[,1] < 158 & X.pred4[,2] > 46) # ~= 0.082
\end{Sinput}
Thus, we are able to predict (based on $\ngen=1000$) that,
for a six-year-old male, the probability of him having a height of $>116$ and a weight of $<21$ is about 10\%;
for a 10-year-old female, the probability of her having a height of $>116$ and a weight of $<21$ is about 25\%;
for a 43-year-old male, the probability of him having a height of $<158$ and a weight of $>46$ is about 12\%; and finally,
for a 67-year-old female, the probability of her having a height of $<158$ and a weight of $>46$ is about 8\%.

\begin{figure}[htbp]
  \centering
  \includegraphics[width=0.49\textwidth]{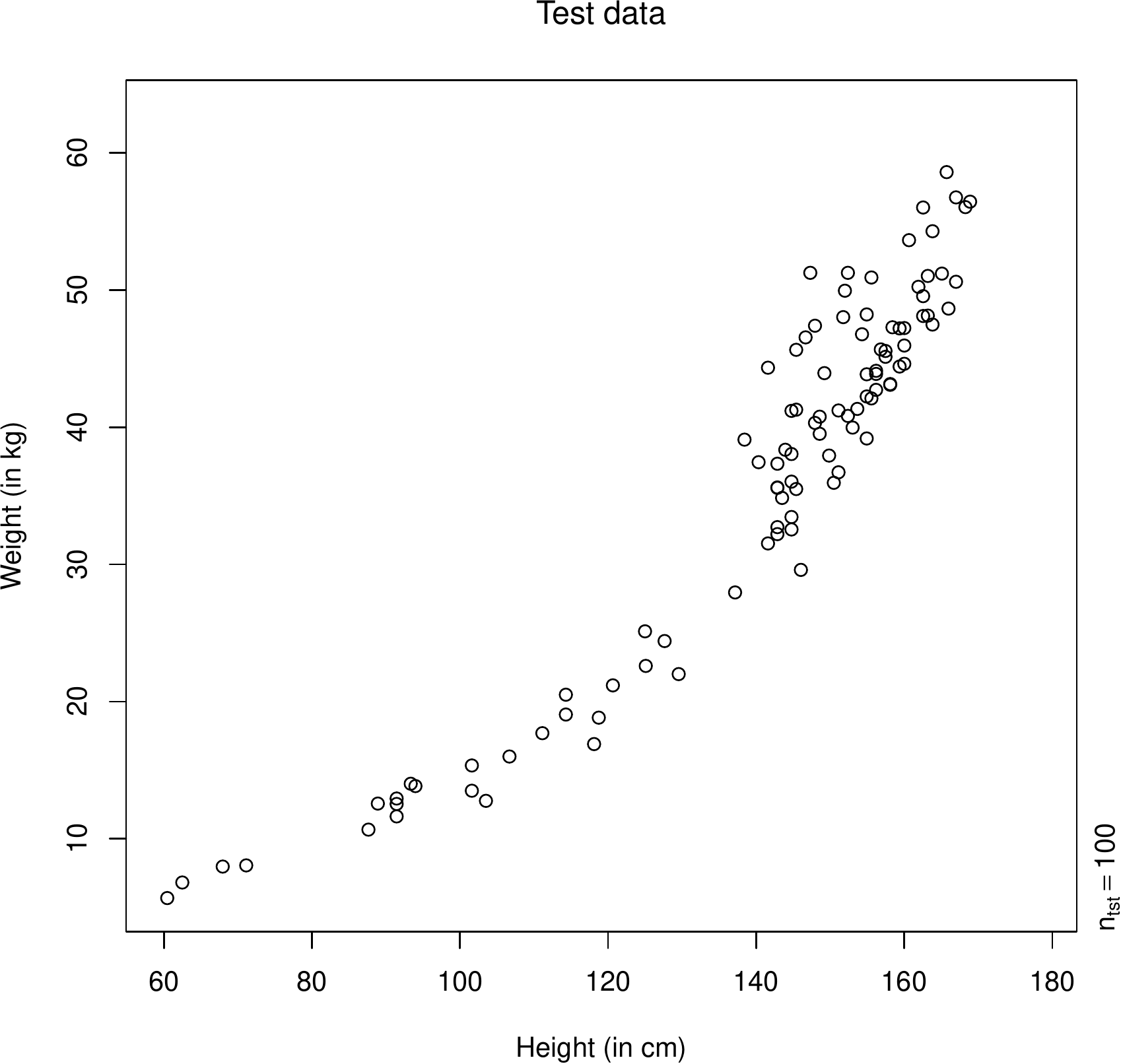}
  \hfill
  \includegraphics[width=0.49\textwidth]{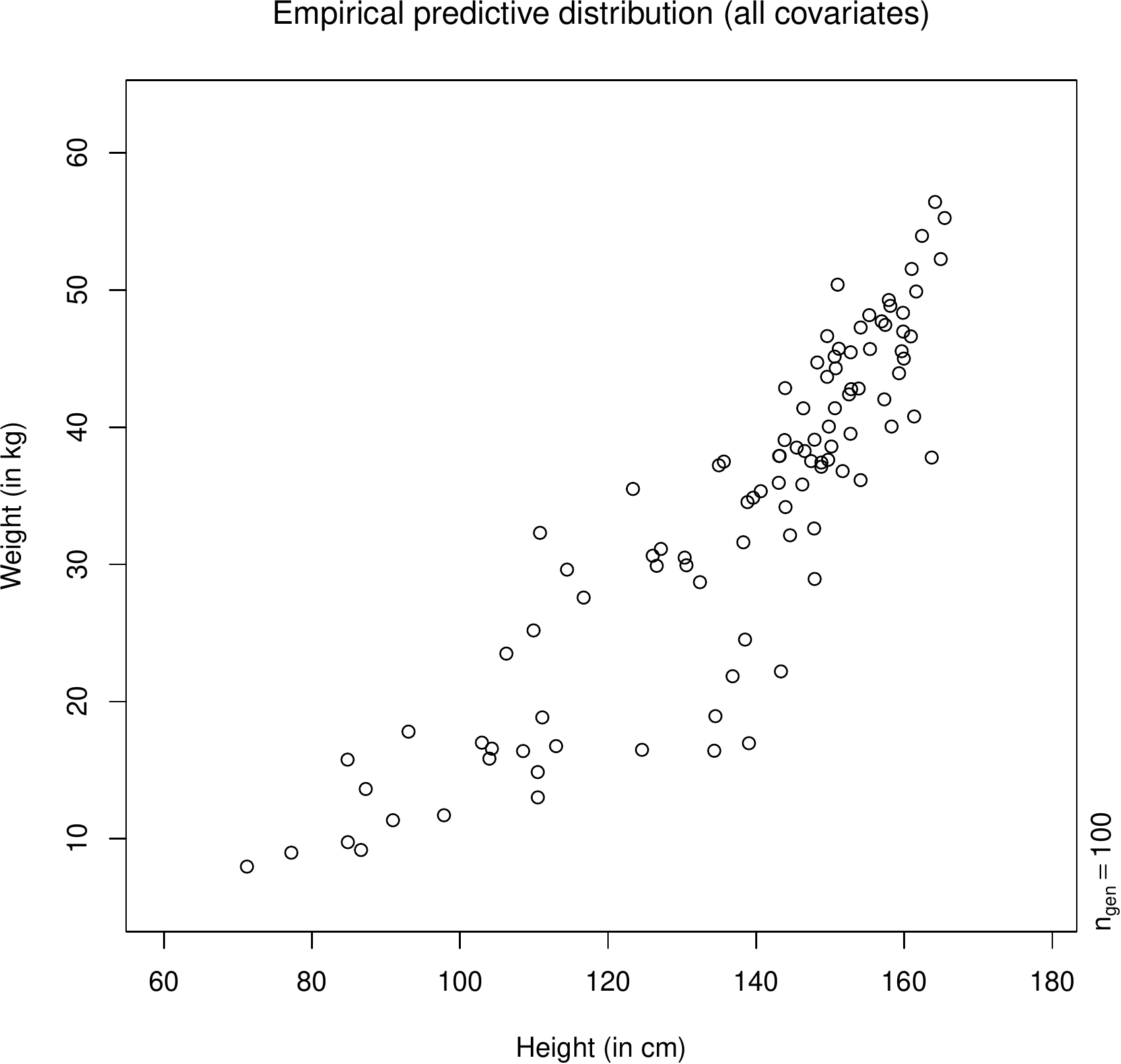}\\[3mm]
  \includegraphics[width=0.49\textwidth]{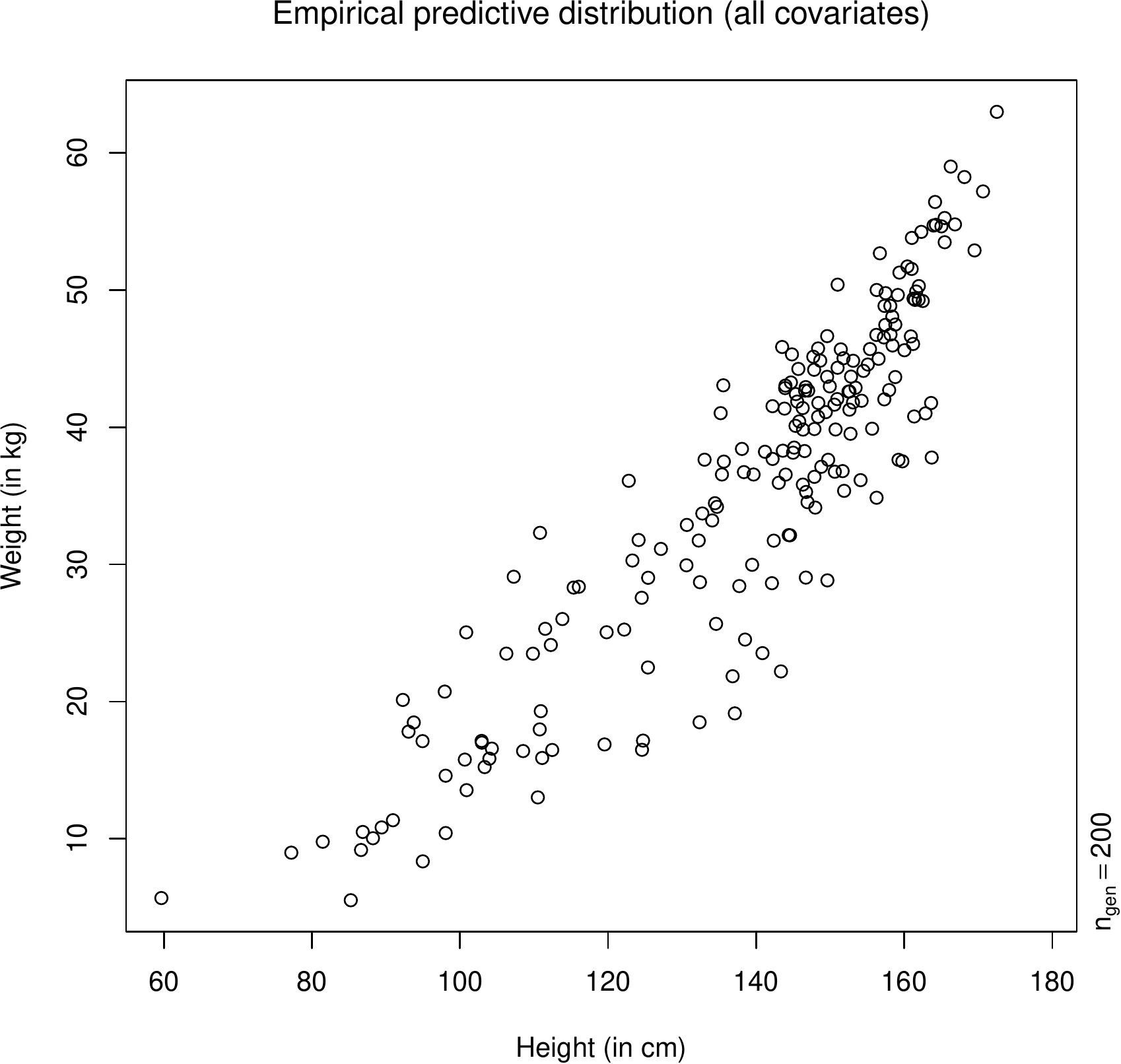}
  \hfill
  \includegraphics[width=0.49\textwidth]{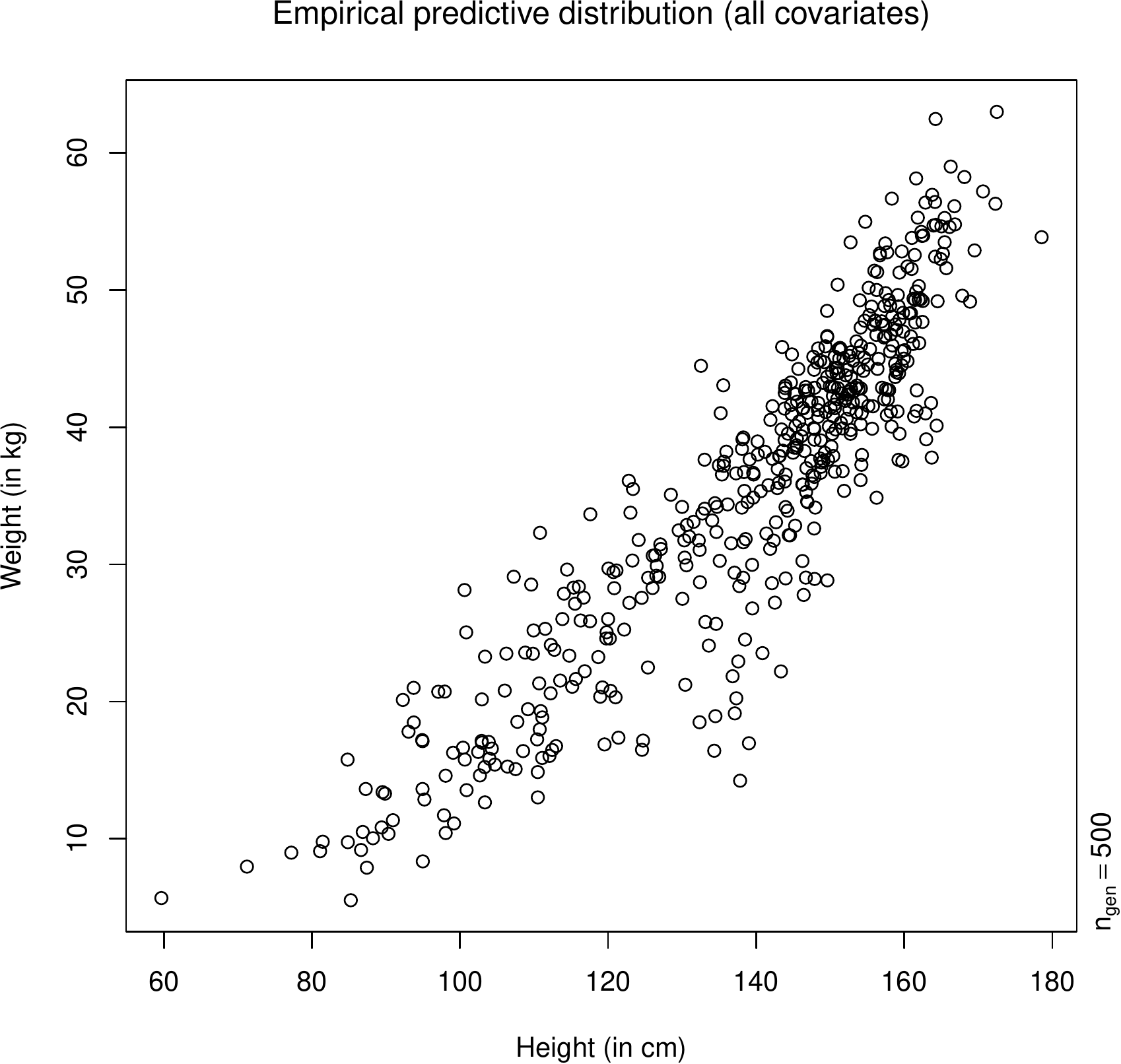}
  \caption{Scatter plot of height and weight data from the test sample (top
    left) and scatter plots of empirical predictive distributions produced by
    the RafterNet where all covariates in the test sample are used and the
    empirical predictive distributions use one (top right), two (bottom left)
    and five (bottom right) samples per given
    covariate.}\label{fig:htwt:pred:full}
\end{figure}

Finally, we use our RafterNet to make probabilistic predictions of height and
weight for all given covariates in the test sample, $\bm{z}_k$,
$k=1,\dots,\ntst$. As before, we follow the procedure in
Algorithm~\ref{tab:rafternet_predict}, but now generate five samples for each
given $\bm{z}_k$ as we iterate over all the $\ntst=100$ observations.
\begin{Sinput}
## Generated samples from each test observation
n.gen.each <- 5 # number of samples to generate for each test observation
U.gen <- rGNN(NN, size = n.tst * n.gen.each, pobs = TRUE) # generate from NN
R.pred <- toEmpMargins(U.gen, x = R.hat) # corresponding residuals
## Create a list of length n.tst containing the indices for test observation i
block <- split(1:(n.tst * n.gen.each), f = rep(1:n.tst, each = n.gen.each))
## Probabilistic prediction using RafterNet for all given covariates in the test data
X.pred <- sapply(1:d, function(j) { # iterate over margins
    sapply(1:n.tst, function(i) { # iterate over each test observation
        predict(raft.fits[[j]], newdata = z.tst[i,]) + # prediction for that observation
        R.pred[block[[i]], j] # residuals
    })
})
\end{Sinput}
Besides five samples for each $\bm{z}_k$, we repeat the algorithm outlined in the
code above to also construct probabilistic predictions based on one and two samples for
each given $\bm{z}_k$. Figure~\ref{fig:htwt:pred:full} displays a scatter plot
of the test data of the height and weight dataset in the top panel, along with
scatter plots of the three probabilistic predictions considered in the bottom
panel. From these plots, we see that the probabilistic predictions essentially
match the height and weight test data.

\end{document}

%
%
%
%

%%% Local Variables:
%%% mode: latex
%%% TeX-master: t
%%% End: